\documentclass[12pt]{article}
\usepackage{graphicx}
\usepackage{amsmath,amsthm,amssymb,bm,subfigure,color,url}
\usepackage[colorlinks=true,citecolor=blue,linkcolor=blue,urlcolor=blue]{hyperref}
\begin{document}

\baselineskip 0.7cm

\begin{titlepage}
\renewcommand{\thefootnote}{\fnsymbol{footnote}}
\begin{flushright}
\end{flushright}

\vskip 1.35cm
\begin{center}
{\large \bf
A Stochastic Temporal Model of\\ Polyphonic MIDI Performance with Ornaments
}
\vskip 1.2cm
Eita Nakamura$^1$\footnote[1]{Electronic address: \tt{eita.nakamura@gmail.com}},
Nobutaka Ono$^1$,
Shigeki Sagayama$^1$   
and Kenji Watanabe$^2$
\vskip 0.4cm

{\it
$^1$ National Institute of Informatics,\\
Tokyo 101-8430, Japan\\
$^2$ Tokyo University of the Arts,\\
Tokyo 110-8714, Japan
}

\vskip 1.5cm
\renewcommand{\thefootnote}{\arabic{footnote}}
\abstract{
We study indeterminacies in realization of ornaments and how they can be incorporated in a stochastic performance model  applicable for music information processing such as score-performance matching.
We point out the importance of temporal information, and propose a hidden Markov model which describes it explicitly and represents ornaments with several state types.
Following a review of the indeterminacies, they are carefully incorporated into the model through its topology and parameters,
and the state construction for quite general polyphonic scores is explained in detail.
By analyzing piano performance data, we find significant overlaps in inter-onset-interval distributions of chordal notes, ornaments, and inter-chord events, and the data is used to determine details of the model.
The model is applied for score following and offline score-performance matching, yielding highly accurate matching for performances with many ornaments and relatively frequent errors, repeats, and skips.
}

\vskip 1cm

{{\bf Keywords:}
stochastic performance model;
ornaments;
hidden Markov model;
score-performance matching;
score following;
performance analysis;
}

\end{center}
\end{titlepage}

\setcounter{page}{2}
\baselineskip 0.6cm

%%%%%%%%%%%%%%%%%%%%%%%%%%%%%%%%%%%%%%%%%%%%%%
\section{Introduction}\label{sec:Intro}
%%%%%%%%%%%%%%%%%%%%%%%%%%%%%%%%%%%%%%%%%%%%%%

Music performance is one of the most important aspects of music and to quantitatively understand how performances are realized and controlled is a fundamental problem in music research.
For this purpose, analysis and quantitative modeling of music performance have been important domains of research in musicology, behavioral science, and music information.
Particularly in music information processing such as score-performance matching (including score following), generation or rendering of expressive performance, music transcription, and rhythm quantization, stochastic models of performance are widely used to derive a set of (often implicit) complex rules that are necessary to construct algorithms for the applications.

In quantitative models of performance, it is essential to describe its indeterminacies and uncertainties properly.
They are included in tempo (both global tempo and tempo variations), noise in onset times, dynamics, and articulations, and also in the way of making performance errors, repeats, and skips, especially in performances during practice \cite{Boulez1963,Palmer1997,Nakamura2014}.
 Hidden Markov model (HMM) among other models is widely used in music information to describe these indeterminacies and uncertainties since it effectively describes the sequential regularities and deformations in music performance together with erroneous and noisy observations, and there are computationally efficient inference algorithms.
 It is successfully applied to the above mentioned tasks \cite{Raphael1999,Cano1999,Otsuki2002,Schwarz2004,Pardo2005,Grindlay2006,Cont2010,Nakamura2014}.

The aim of this paper is to discuss ornaments, which are yet another major source of indeterminacies in music performance.
In addition to their importance in music expression in Western classical music, their improvisational nature provides interesting problems and challenges in music research.
Indeterminacies of ornaments have been studied in Refs.~\cite{Neumann1983,Desain1994,Repp1997,Windsor2001} in view of musicology and behavioral science with some interesting quantitative analyses, and their relevance in music information processing is discussed in Refs.~\cite{Dannenberg1988,Schwarz2004,Cont2010,Gingras2011}.
Given the musical interests and applicational need, it is worthwhile to study how to incorporate indeterminacies of ornaments into a stochastic model of performance that is applicable to music information processing.

As an explicit application, we discuss score-performance matching, both real-time online matching (a.k.a.~score following) and offline alignment, which is a popular field of research \cite{Dannenberg1984,Vercoe1984,Desain1997,Heijink2000,Orio2003,Pardo2005,Tekin2006,Cont2010,Gingras2011,Arzt2012,Nakamura2014} and one of the most basic techniques for music information processing and performance analysis.
Since the indeterminate nature of ornaments can cause troubles in recognizing the score position, the significance of treating ornaments in score-performance matching has been indicated repeatedly \cite{Dannenberg1988,Cont2010,Gingras2011,Nakamura2014}.
A method using preprocessing is proposed in Ref.~\cite{Dannenberg1988}, but it can fail under performance errors as mentioned in the paper and also has trouble in unexpected situations such as repeats and skips, which motivates the use of a stochastic method.
In Ref.~\cite{Schwarz2004}, the idea of representing a trill as a state in HMM is mentioned, but an explicit realization of the model is not given.
For audio signals, a hidden hybrid Markov/semi-Markov model is proposed to describe performances with ornaments, where trill, short appoggiatura, and glissando are described as special states \cite{Cont2010}.
Since we need quite different treatment for audio signals and symbolic signals \cite{Nakamura2014}, online and offline algorithms based on stochastic method are also desired for symbolic performance signals in musical instrument digital interface (MIDI) format, which are used in performance analyses and also important in technological applications.
Although concurrent ornaments in polyphonic passages are discussed in Ref.~\cite{Gingras2011}, systematic discussion or evaluation on complex cases including a concurrence of an arpeggio and a trill, which appear for example in the pieces of Liszt and Chopin, has not been given in the literature.
Given the fact that multiple ornaments can overlap or appear simultaneously in polyphony, an extensive discussion on the general case alongside of the problem of score representation is in order.

In this paper, we propose an HMM for polyphonic MIDI performance with ornaments.
We discuss in detail the indeterminacies in most important ornaments, particularly focusing on their relevance in complex polyphonic passages and the issue of computational score representation.
As we will discuss in Sec.~\ref{sec:Indeterminacies}, temporal information is crucial when dealing with ornaments, and we also confirm this fact quantitatively by performance analysis in Sec.~\ref{sec:Analysis}.
The temporal information is explicitly described as an additional dimension in the state space, and we show that the model is equivalent to an HMM that outputs inter-onset interval (IOI), which is similar to models in Refs.~\cite{Cemgil2000B,Otsuki2002}.
The present performance model is an extension of the model proposed in Ref.~\cite{Nakamura2014}, and ornaments are described with additional types of states.
It accommodates performance errors, arbitrary repeats and skips without serious increase in computational cost.
The construction of the state sequence from a given score is carefully derived and explained in detail.

Results of analyzing piano performance data are presented and used to determine details of the model and to fix its parameters.
We construct score-performance matching algorithms from the proposed model and explain their advantages and disadvantages in comparison with other algorithms.
In general, our algorithms have advantages in computational efficiency and they can handle arbitrary repeats and skips in performances.
The algorithms are evaluated and compared to other algorithms as far as possible.
Finally we summarize and discuss prospective issues in stochastic modeling of performance and possible other applications of the present model.
We are willing to share our algorithms and evaluation data for future studies, and contacts are welcome to the corresponding author in this regard.

%%%%%%%%%%%%%%%%%%%%%%%%%%%%%%%%%%%%%%%%%%%%%%
\section{Indeterminacies in realization of ornaments}\label{sec:Indeterminacies}
%%%%%%%%%%%%%%%%%%%%%%%%%%%%%%%%%%%%%%%%%%%%%%

\subsection{Types of ornaments and their indeterminacies}\label{sec:OrnamentsAndIndeterminacies}

In this paper, we mainly consider Western classical music during the common practice period, that is, from the late baroque period to early twentieth century, although this does not mean that the discussion can only be applied to the particular music.
Music in the period is written in the common metric notation system, in which scores basically describe movements of musical instruments or actions of performers required to realize the music.
Performances based on these scores generically have indeterminacies and uncertainties described in the Introduction due to ambiguities and indeterminate nature of the score, performer's skill, and physical constraints of musical instruments \cite{Boulez1963,Palmer1997}.
Ornaments are another major source of indeterminacies and uncertainties.

To begin our discussion on ornaments, we first define the scope of the discussion and what is here meant by ``ornaments''.
In general, ornaments are divided into notated and improvised (or free) ornaments.
Both improvised ornaments and performance errors can introduce notes into a performance which have no corresponding symbols in the score.
While listeners can generally distinguish ornaments from errors, and it is important to do so in situations such as performance error analysis, they are treated similarly in our score-performance matching algorithm described below.
(See Ref.~\cite{Gingras2011} for a discussion on identifying performance errors and ornaments.)
Our focus here is on how to model the performance of notated ornaments.

In chapter 13 of Ref.~\cite{Read1969}, commonly used ornaments are listed; trill, tremolo, short appoggiaturas\footnote{This is written as ``grace notes'' in Ref.~\cite{Read1969}. In general, the word ``grace notes'' means either small notes in scores or ornamental figures notated with these small notes, which are also called short appoggiaturas. We will use the word ``short appoggiatura'' to mean the ornamental figure and ``grace note'' to mean a small note in scores in this paper, to avoid confusion.}, long appoggiatura\footnote{Unlike short appoggiaturas, long appoggiaturas (or simply, appoggiaturas) usually have determinate note values. Typically they are notated with a single grace note (without a slash), and a single short appoggiatura is usually notated with a grace note with a slash.}, arpeggio, glissando, mordent, and turn.
They are usually notated with special symbols or with grace notes.
There are other types and symbols of ornaments, for example, slide and various combined ornaments, that appeared especially in the baroque period, and their conventions and interpretations are often discussed and associated with different periods, regions, and composers (see e.g., Ref.~\cite{Neumann1983} and the article ``Ornaments'' in Ref.~\cite{NewGrove2001} and references therein).
Our focus is more on the notational ambiguity and interpretive nature of ornaments, rather than their relevance to compositional or aesthetic effects.
In this sense, long appoggiaturas are more a matter of pure notation, and we will not discuss them in the following since they can be almost equivalently notated with usual (not grace-) notes\footnote{It is true that notational ambiguities with long appoggiaturas or confusion between long and short appoggiaturas sometimes arise, but these require more or less musicological arguments which are out of our scope.}.
For definiteness, we confine ourselves to the above listed ornaments in Ref.~\cite{Read1969} other than long appoggiaturas in the following, and other ornaments will only be mentioned when necessary.
How ornaments can be interpreted in modern practice is essential for the discussion, but it is not our aim to study how they are interpreted by particular performers nor how they should be interpreted musicologically.

\begin{table}[tb]
\caption{List of most frequent ornaments and their indeterminacies.}
\label{tab:ListIndeterminacies}
\begin{center}
\begin{tabular}{c|c}\hline\hline
Ornament&Indeterminacies\\
\hline\hline
\shortstack{Trill\\~} & \shortstack{Rapidity; \# of notes; addition of after notes;\\ addition/deletion of the initial upper note}\\\hline
Tremolo & Rapidity\\\hline
Short appoggiatura & Rapidity; relative timing to metrical beat\\\hline
After note & Rapidity\\\hline
\shortstack{Mordent \& turn\\~} & \shortstack{Rapidity; addition of an initial note;\\relative timing to metrical beat}\\\hline
\shortstack{Arpeggio\\~} & \shortstack{Rapidity; overlap between hands; ordering;\\relative timing to metrical beat}\\\hline
Glissando & Rapidity; range (if not specified)\\
\hline\hline
\end{tabular}
\end{center}
\end{table}
Indeterminacies in realization of ornaments derive mostly from their symbolic and interpretive nature (Table \ref{tab:ListIndeterminacies}).
For example, in the realization of a trill, the rapidity and consequently the number of performed notes differ on occasion\footnote{By this, we mean that they may differ between performers and also from time to time.} due to performers' interpretation and skill, and also by chance.
In the case of a long trill, the rapidity can vary in time, often starting with slow alternation and then making it faster.
Other common indeterminacies are the choice of starting with the principal note or the upper note, and of adding short appoggiaturas, usually consisting of the lower note and the principal note, when they are not notated explicitly.
Tremolos have similar indeterminacies.
Tremolos in which each note or chord has a definite note value are called measured, and otherwise they have undetermined rapidity and are called unmeasured \cite{Read1969}.
Due to notational confusion, measured tremolos are sometimes played as unmeasured tremolos, and certain tremolos are not easy to be attributed measured or unmeasured uniquely, resulting in uncertainties of realization in effect.

For short appoggiaturas, the sequence of notes is determined, but temporal indeterminacies exist.
As well as their durations, the timing of their onsets relative to metrical beat is generically indeterminate.
Typically the first note of short appoggiaturas is performed on the beat (accented) or the principal note after them is performed on the beat (unaccented) \cite{Read1969,Turk1789}.
The indeterminacy is more explicit when short appoggiaturas appear in polyphonic passages as the ordering of the notes between the short appoggiaturas and notes in other voices may vary with interpretations.
Sometimes the timing is indicated with a slur or their relative position to a bar line, or it can be implied by context such as the case for the grace notes after a trill.
In case that short appoggiaturas are indicated or (almost) unambiguous to be performed in precedence over the beat, we call them after notes\footnote{In German terminology, accented short appoggiaturas are sometimes called ({\it kurze}) {\it Vorschl\"age} and unaccented ones {\it Nachschl\"age}.}.

\begin{figure}[tbp]
\begin{center}
\subfigure[Upper mordent]
{\includegraphics[clip,width=0.26\columnwidth]{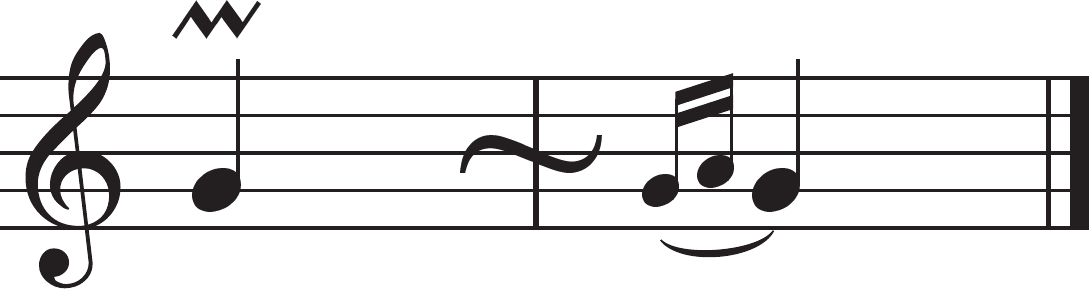}}
\subfigure[Direct turn]
{\includegraphics[clip,width=0.26\columnwidth]{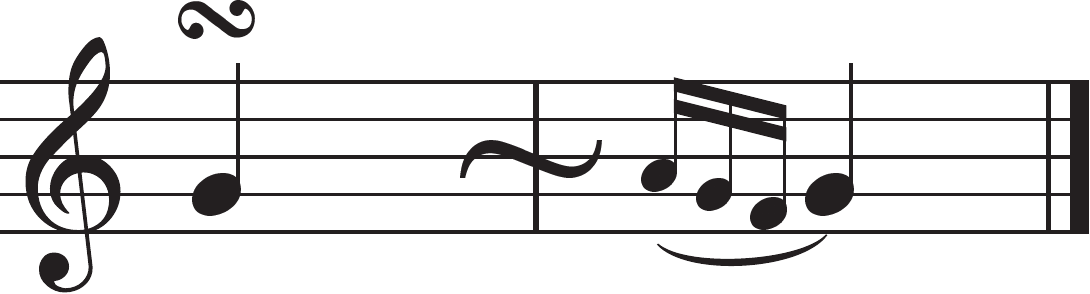}}
\subfigure[Delayed turn]
{\includegraphics[clip,width=0.4\columnwidth]{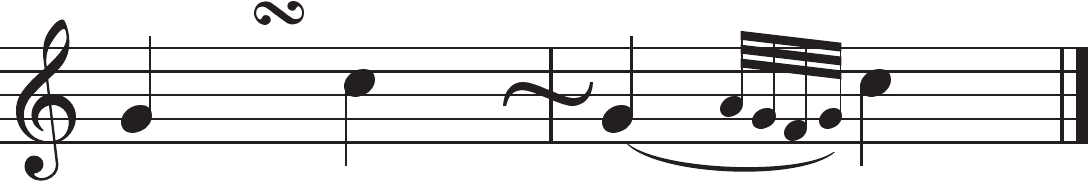}}
\end{center}
\caption{Tentative representations of mordents and turns in terms of short appoggiaturas and after notes.}
\label{fig:TentativeRepresentations}
\end{figure}
The case of the mordent and turn is similar.
In the simplest interpretation, a mordent or a turn can be represented with short appoggiaturas or after notes (Fig.~\ref{fig:TentativeRepresentations}).
The quasi-equivalence of these representations is implied in Ref.~\cite{Turk1789}, and they are alternatively used in musical pieces (e.g., the first movement of Schubert's piano sonata in A minor D.~485; Czerny's Ops.~365-8, 261-79, and 261-80) and in different editions of same pieces (compare, e.g., turns around the first repeat sign in the fifth variation in the first movement of
Mozart's piano sonata in A major K.~331 in different editions\footnote{Copy of the first Artaria edition, Breitkopf \& H\"artel, Peters, and Schirmer editions can be downloaded from IMSLP Petrucci Music Library \url{http://imslp.org}.}).
For an upper mordent (or Pralltriller), there is also a choice of adding the upper note at the head.
Particularly in baroque music, upper and lower mordents are realized with additional alternations, or even as a long trill.
There are two types of turns, direct turn and delayed turn \cite{Read1969}, and their typical interpretations are illustrated in Fig.~\ref{fig:TentativeRepresentations}.
In general, there is a choice to add the principal note at the head of a direct turn.
The rapidity of a turn is also much indeterminate, especially in slow passages.

Arpeggios have similar indeterminacies as a sequence of short appoggiaturas, where the rapidity of rolling and the timing with respect to beat are generally indeterminate.
Arpeggios involving both hands of keyboard playing can either be broken, in which the bottom notes of both hands sound simultaneously ideally, or unbroken, in which the whole chord is rolled as a succession of single notes \cite{Read1969}.
In reality, asynchrony between both hands in a broken arpeggio can be large, and notes played by both hands in an unbroken arpeggio can overlap \cite{Repp1997}, resulting in changes in expected ordering of note onsets.

Glissando can be performed with different speed which can change in time.
Occasionally the range is only partially indicated, which cause intended indeterminacy in the number of notes and rapidity.
For simultaneous multiple glissandos such as an octave glissando, the ordering of notes across voice can be different from the ideal realization.
Similarly, a relative timing and ordering of notes between glissando and other voices are generally uncertain.

Finally, in polyphonic passages, the ornaments can simultaneously appear in different voices and the above indeterminacies are superposed.
We have already mentioned several such effects in the above.
Another typical example is a double (or triple) trill, which can involve a single hand or both hands.
A double trill is typically played in almost synchrony or in simple integral ratios, but the synchronization may become loose for fast trills.

\subsection{Significance for score-performance matching}\label{sec:SignificanceForMatching}

Given the indeterminacies in ornaments described in the previous section, one must treat ornaments with care in music information processing.
As an explicit example, we consider score-performance matching.
For trills and unmeasured tremolos, it is not meaningful to match performed notes to a particular set of explicitly realized notes.
For trills, a successful matching algorithm must correctly treat addition (or deletion) of the upper note at the head and after notes in the end.
To match short appoggiaturas or arpeggios in polyphonic passages correctly, the algorithm should hold rules consistent with indeterminacy in local ordering of notes.
Similar case is for mordent and turn.

Another problem arises in clustering of notes. Suppose a passage in which a chord is repeated several times.
Local ordering of chordal notes is generally indeterminate due to noise in onset times.
If note deletions and insertions happen, one must use temporal information such as IOI to match the notes unambiguously.
Use of a threshold on IOI works well in this case since the distribution of IOI between chordal notes has little overlap with that of IOI between notes in adjacent chords (inter-chord IOI) \cite{Bloch1985,Nakamura2014}.
In contrast, as we will confirm quantitatively in Sec.~\ref{sec:Analysis}, IOIs involving short appoggiaturas and arpeggios can be as large as inter-chord IOIs, and the clustering is less trivial.
The same problem arises in upper mordents and direct turns due to indeterminate addition of an initial note.
Therefore the use of temporal information is essential for performances with ornaments.

To solve these problems, a preprocessing method for handling trill and glissando in online matching is proposed in Ref.~\cite{Dannenberg1988}.
The idea is to preprocess performed notes so that ornamental notes are not sent to the matching module directly.
It possibly works because we can anticipate ornaments in the score from score-position estimation.
However, as is mentioned in the reference, the preprocessing can fail when there are performance errors, for instance, when a note just before an ornament is omitted.
Also, in light of allowing arbitrary repeats and skips \cite{Nakamura2014}, there is additional risk in using the preprocessor depending heavily on anticipations, since repeats and skips can hardly be anticipated.
It is not easy to apply the preprocessing method to various ornaments in highly polyphonic passages and to offline matching.

For offline matching, a method of identifying ornaments based on perceptual principles is proposed in Ref.~\cite{Gingras2011}, in which pitch, temporal, and voice informations are used.
The method is general and applicable for both notated and improvised ornaments.
The matching technique cannot be applied to performances with large repeats and skips directly, although it may be possible in principle.

Another way is to build a stochastic model of performance which can properly describe the indeterminacies and uncertainties, as is the aim of this paper, and use it for constructing a matching algorithm.
Generally, use of a stochastic model has advantages in organizing complex rules without inconsistencies or conflicts and setting model parameters in a principled way such as the maximal likelihood method.
Additional bonus of using HMM here is that one can obtain both online- and offline matching algorithms simultaneously.
There have been attempts to incorporate ornaments into HMM \cite{Schwarz2004,Cont2010}, but a fully appropriate model for polyphonic MIDI performance has not been proposed, as explained in Sec.~\ref{sec:Intro}.
Our model based on HMM will be presented in Sec.~\ref{sec:PerformanceModel} after describing score representation of polyphonic music with ornaments in the next section.

\subsection{Score representation}

In order to systematically study ornaments and to show the generality and limitation of the discussion, we clarify the definition and representation of scores.
We define score as a polyphonic passage, which is composed of one or more ``voice parts''.
Each ``voice part'' is a linear sequence of musical events; ``chords''\footnote{In this paper, the term ``chord'' will be used in a way which is different from its normal meaning. We define the term in the next sentence.}, rests, tremolos, and glissandos.
Here a ``chord'' consists of one or more notes whose onsets and offsets are notated as synchronous on the score.
Notes in a chord can be ornamented as trill, upper and lower mordent, direct and delayed turn (normal and inverted), and other embellishments typical of the Baroque period, such as the slide and the double-cadence \cite{Neumann1983}, that will not be discussed in detail but can be treated similarly.
It is specified by constituent pitches and a note value, together with ornamentation information.
A rest is specified by a note value.
A tremolo is specified by a set of chords and a note value. We here consider unmeasured tremolos, and definitive measured tremolos can be described as a sequence of chords.
A glissando is typically specified by start tone(s), end tone(s), a scale, and a note value indicating the duration of the glissando, and occasionally the range of tones is not fully specified.
We restrict ourselves to the case where the range is specified since this is almost all the case for music in the common practice period, and other cases might be treated similarly\footnote{For example, we might take the range of glissando sufficiently wide.}.

In a voice part, each of these events can be preceded by short appoggiaturas and succeeded by after notes, both of which can be a sequence of chords in general.
Generically, these chords are notated as grace notes and their durations are not metrically specified.
Since the so-called long appoggiaturas are notational convention and can usually be replaced by ordinary chords, we treat them as chord events.
In summary, a voice part $H$ is written as
\begin{equation}\label{eq:VoicePart}
H=\alpha_1\beta_1y_1 \cdots \alpha_n\beta_ny_n,
\end{equation}
where $y_i$ is either a chord, a rest, a tremolo, or a glissando, and $\alpha_i$ and $\beta_i$ denotes after notes and short appoggiaturas, which can be empty if there is none.
The factor $y_i$ is said empty if it is a rest.
Note that in the convention, $\alpha_i$, $\beta_i$, and $y_i$ have the same score time.
A fermata may be put upon a chord, a rest, or a tremolo.
A notated cadenza is a sequence of chords typically associated with a fermata and notated with grace notes.
We can describe these indications as additional data on musical events in Eq.~(\ref{eq:VoicePart}).

A polyphonic passage $\bm H$ composed of a set of voice parts $H_1,\cdots,H_V$ is denoted by
\begin{equation}\label{eq:PolyphonicPassage}
\bm H=\bigoplus_{v=1}^VH_v,
\end{equation}
where each $H_v$ has the form of Eq.~(\ref{eq:VoicePart}) and we have used a direct sum symbol to indicate a composition of voice parts ($V$ is the number of voice parts).
An arpeggio is an indication of rolling notes that have simultaneous score time, typically from lower pitches to higher pitches.
It may involve several voice parts (e.g.~Chopin: \'Etude Op.~10-8, bar 79 \cite{Chopin1879}) and short appoggiaturas (e.g.~Chopin: \'Etude Op.~10-11, bar 34; Op.~25-5, bar 43 \cite{Chopin1879}), and multiple arpeggios can occur simultaneously (e.g.~Chopin: \'Etude Op.~10-11 \cite{Chopin1879}).
In our score representation, an arpeggio is specified as a subset of notes in $\bm H$ with simultaneous score time, possibly with an indication for ordering, typically up or down.
An example of the score representation will be given in Fig.~\ref{fig:ExScoreRep}.

We cannot assure that the score representation is general enough to cover all pieces in the common practice period, but we empirically checked that exceptions out of the score representation are at least very rare.
The representation is compatible with the MusicXML format, a common sheet music notation format (\url{http://www.musicxml.com}), except that after notes and short appoggiaturas are not distinguished within the notation per se.

%%%%%%%%%%%%%%%%%%%%%%%%%%%%%%%%%%%%%%%%%%%%%%
\section{Performance model}\label{sec:PerformanceModel}
%%%%%%%%%%%%%%%%%%%%%%%%%%%%%%%%%%%%%%%%%%%%%%

\subsection{Temporal HMM and IOI output}

In the following, we extend the model in Ref.~\cite{Nakamura2014} to incorporate temporal information.
The state space of the current model is represented by a pair $(i_m,t_m)$ of intended musical event $i_m$ and onset time $t_m$.
Here, $i$ labels musical events in the performance score, which are described in detail below, and $m=1,\cdots,M$ indexes the performed notes with the total number $M$.
The probability of occurrence of $(i_m,t_m)$ is in general dependent of the previous performed events, and an approximate model is obtained by assuming that the dependence is Markovian.
With the assumption of time translational invariance, the state transition probability is given as
\begin{equation}
P(i_m,t_m|i_{m-1},t_{m-1})=a'_{i_{m-1},i_m}(t_m-t_{m-1})
\end{equation}
where $a'$ is some function with a normalization condition
\begin{equation}\label{eq:NormalizationTrProbFormer}
\sum_{i_m}\int_0^\infty ds\, a'_{i_{m-1},i_m}(s)=1.
\end{equation}
What we actually observe is a performed pitch, not the intended event, and it is also stochastically described.
Assuming that the observation process is dependent only on the current and previous states, the output probability can be written as
\begin{equation}
P(p_m|i_{m-1},t_{m-1};i_m,t_m)=b'_{i_{m-1},i_m}(p_m;t_m-t_{m-1}).
\end{equation}
Here $b'$ is some function satisfying
\begin{equation}\label{eq:NormalizationOutProbFormer}
\sum_{p_m}b'_{i_{m-1},i_m}(p_m;t_m-t_{m-1})=1,
\end{equation}
where $p_m$ denotes the pitch of the $m$-th performed note.
Combining these probabilities, the probability of the sequence of performance $(p_m,i_m,t_m)_{m=1}^M$ is given as
\begin{equation}\label{eq:PerformanceProbFormer}
P\left((p_m,i_m,t_m)_{m=1}^M\right)=\prod_{m=1}^M
a'_{i_{m-1},i_m}(t_m-t_{m-1})b'_{i_{m-1},i_m}(p_m;t_m-t_{m-1}),
\end{equation}
where, by abuse of notation, the factors for $m=1$ mean the initial probabilities.

In the above model, onset time is described as a dimension in the state space.
Since the onset time $t_m$ and the IOI $\delta t_m=t_{m}-t_{m-1}$ are observables, we can also regard these temporal quantities as generated by corresponding transitions between musical events.
We can show that these two views are indeed equivalent.
By defining
\begin{align}
a_{i_{m-1},i_m}&=\int_0^\infty ds\,a'_{i_{m-1},i_m}(s),\label{eq:TrProb}\\
b_{i_{m-1},i_m}(p_m,\delta t_m)
&=\frac{a'_{i_{m-1},i_m}(\delta t_m)b'_{i_{m-1},i_m}(p_m;\delta t_m)}{a_{i_{m-1},i_m}},
\label{eq:OutProb}
\end{align}
Eq.~(\ref{eq:PerformanceProbFormer}) can be rewritten as
\begin{equation}\label{eq:PerformanceProbLatter}
P\left((p_m,i_m,t_m)_{m=1}^M\right)=\prod_{m=1}^M
a_{i_{m-1},i_m}b_{i_{m-1},i_m}(p_m,\delta t_m).
\end{equation}
We can interpret $a_{i_{m-1},i_m}$ and $b_{i_{m-1},i_m}(p_m,\delta t_m)$ as the probability of transition from $i_{m-1}$ to $i_m$ and the output probability of a pair of observations $(p_m,\delta t_m)$ resulting from the transition.
Note that the normalization conditions in Eqs.~(\ref{eq:NormalizationTrProbFormer}) and (\ref{eq:NormalizationOutProbFormer}) yield normalizations for the new probabilities properly as
\begin{equation}
\sum_{i_m}a_{i_{m-1},i_m}=1\quad{\rm and}\quad
\sum_{p_m}\int_0^\infty ds\,b_{i_{m-1},i_m}(p_m,s)=1.
\end{equation}
It is easy to see that the original probabilities $a'_{i_{m-1},i_m}(\delta t_m)$ and $b'_{i_{m-1},i_m}(p_m;\delta t_m)$ can be reproduced from $a_{i_{m-1},i_m}$ and $b_{i_{m-1},i_m}(p_m,\delta t_m)$, and hence the two models are equivalent.

The current model is an HMM which extends the model in Ref.~\cite{Nakamura2014} with an additional dimension of time in the state space, or with an output of IOI.
In what follows, we describe the performance model in terms of the HMM with IOI output, and $i$ indexes HMM states corresponding to musical events.
For early applications in music information of similar stochastic models involving onset times or IOIs, see Refs.~\cite{Saito1999,Large1999,Cemgil2000,Otsuki2002}.

The model parameters in Eqs.~(\ref{eq:TrProb}) and (\ref{eq:OutProb}) are to be fitted to the actual performance data. 
However, it is hard to obtain sufficient amount of data to set the output probability $b_{ij}(p,\delta t)$ directly.
We compromise on the problem by assuming that it is factorized into two independent output probabilities, one describes the distribution of pitch and the other IOI.
The assumption yields another advantage of low computational cost.
It is further assumed that the output probability of pitches is only dependent on the current state for simplicity.
Thus, the output probability is written as $b_{ij}(p,\delta t)=b_j^{\rm pitch}(p)b_{ij}^{\rm IOI}(\delta t)$.

\subsection{State construction by hierarchical model}\label{sec:StateConstruction}

To represent music performance by the HMM, one should relate music events in score to states in the model.
In general, there are several possibilities.
For example, a chord can be represented as a state, and attacks of multiple notes in the chord can be described as self transitions with the output probability nearly equally distributed for all chordal pitches as in Ref.~\cite{Nakamura2014}.
One can also represent a chord with multiple states, each corresponding to a note in the chord, and the output probability is high for the pitch of the note.
Randomly ordered attacks of notes in the chord can then be described as mixed transitions within the multiple states.
In the latter representation, one can, for example, describe the structure of internal transitions within a chord, and the descriptive power is in general stronger, but efficiency in computation and parameter fitting is then worse since there are more states and parameters.

\begin{figure}[tbp]
\begin{center}
\includegraphics[clip,width=0.6\columnwidth]{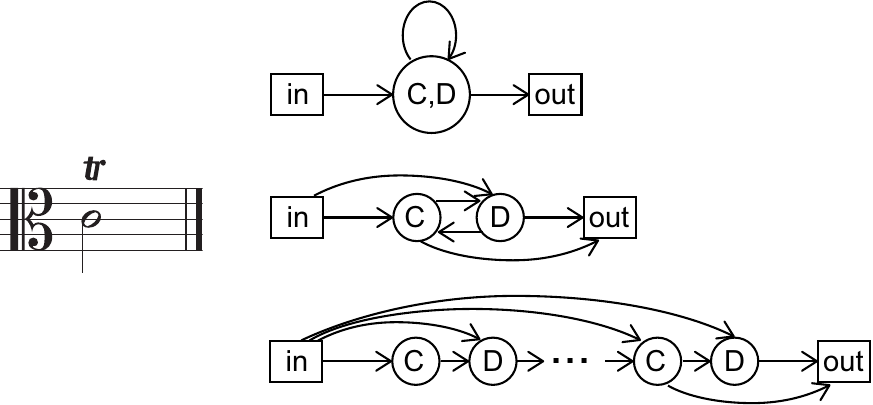}
\end{center}
\caption{Examples of state representation of a trill.}\label{fig:TrillRepresentation}
\end{figure}
Another example is representation of a one-note trill (Fig.~\ref{fig:TrillRepresentation}).
It can be represented as a state, as two states which correspond to the principal note and the upper note, or as a chain of states whose length can stochastically describe the number of performed notes similarly as the variable duration model \cite{Ferguson1980}.
There is generically a trade-off between simplicity/efficiency and complexity/preciseness.

In a general setting, the model is concisely described as a two-level hierarchical model \cite{Fine1998}, in which a state in the top level corresponds to a musical event.
The HMMs in the two levels will be called top- and bottom HMM.
The hierarchical HMM can be expanded into an ordinary HMM, and the bottom-level states are in one-to-one correspondence to states in the expanded HMM.
Let $A_{IJ}$ denote the transition probability from state $I$ to $J$ in the top level, and let $\rho^{(I)}_{k\ell}$ denote the transition probability in the bottom level from substate $k$ to $\ell$ of state $I$.
The entering and exiting probabilities of substate $k$ are denoted by $\rho^{(I)}_{{\rm in},k}$ and $\rho^{(I)}_{k,{\rm out}}$, satisfying $\sum_k\rho^{(I)}_{{\rm in},k}=1$ and $\sum_\ell\rho^{(I)}_{k\ell}+\rho^{(I)}_{k,{\rm out}}=1$ for all $k$.
The transition probability of the expanded HMM from state $i=(I,k)$ to $j=(J,\ell)$ is given as
\begin{equation}\label{eq:HierarchicalTrProb}
a_{ij}=a_{(I,k)(J,\ell)}=\begin{cases}
\rho^{(I)}_{k,{\rm out}}A_{IJ}\rho^{(J)}_{{\rm in},\ell}, &\text{if $I\neq J$};\\
\rho^{(I)}_{k\ell}+\rho^{(I)}_{k,{\rm out}}A_{II}\rho^{(I)}_{{\rm in},\ell}, &\text{if $I=J$}.
\end{cases}
\end{equation}
The $A_{IJ}$ corresponds to the event-level transition probability, and it describes straight transitions to the next state, insertions and deletions of events, and large repeats and skips, similarly as the chord-level transition probability in Ref.~\cite{Nakamura2014}.
Because the output probability of our model is of Mealy type, which means that it depends on both the current and previous states, we will discuss it later with a little care.

In the following we consider one of the simplest realizations of the model concretely.
For this, we first consider a generalization of Conklin's ``homophonization'' \cite{Conklin2002}.
Given a score $\bm H$ in Eq.~(\ref{eq:PolyphonicPassage}), we construct a linear sequence (called the homophonization $\tilde{\bm H}$ of $\bm H$)
\begin{equation}\label{eq:HomophonizedSeq}
\tilde{\bm H}=\tilde{\bm\alpha}_1\tilde{\bm\beta}_1\tilde{\bm y}_1
\cdots \tilde{\bm\alpha}_N\tilde{\bm\beta}_N\tilde{\bm y}_N,
\end{equation}
where the symbols $\tilde{\bm\alpha}_I$, $\tilde{\bm\beta}_I$, and $\tilde{\bm y}_I$ ($I=1,\cdots,N$) are composites of after notes, of short appoggiaturas, and of measured notes at some score time $\tilde{\tau}_I$, written as
\begin{equation}\label{eq:ComponentsOfHomophonization}
\tilde{\bm\alpha}_I=\bigoplus_v\tilde{\alpha}_{I,v},\quad
\tilde{\bm\beta}_I=\bigoplus_v\tilde{\beta}_{I,v},\quad
\tilde{\bm y}_I=\bigoplus_v\tilde{y}_{I,v}.
\end{equation}
Here a unit corresponding to each $I$ in $\tilde{\bm H}$ is constructed if there happens new structure in onset events in $\bm H$ at $\tilde{\tau}_I$.
At the stage of homophonization, upper and lower mordents and turns are transformed to short appoggiaturas and after notes as in the representation in Fig.~\ref{fig:TentativeRepresentations}, and glissandos are expanded into ordinary notes.

A composite factor is said to be empty if all of its component factors are empty.
We assume that at least one of the factors of $\tilde{\bm\alpha}_I$, $\tilde{\bm\beta}_I$, or $\tilde{\bm y}_I$ is not empty and there are no redundancies in the representation in Eq.~(\ref{eq:HomophonizedSeq}).
Especially we have $\tilde{\tau}_I\neq\tilde{\tau}_{I'}$ if $I\neq I'$.
We also define $\tilde{\tau}^{\rm end}_I$ as the score time after which no new onsets that are part of $\tilde{\bm y}_I$ can occur.
Details and an algorithmic construction of the homophonization are described in Appendix \ref{sec:Homophonization}.
$\tilde{\bm H}$ is associated to the state sequence of the upper-level HMM.
We take a factor in $\tilde{\bm H}$ within a score time, i.e., $\tilde{\bm\alpha}_I\tilde{\bm\beta}_I\tilde{\bm y}_I$, as a state in the upper-level HMM.

\subsection{Event model}\label{sec:EventModel}

Let us now explain the bottom HMM, or the event model.
As units of bottom-level state, we take the minimum units of score notes that are well-ordered in ``straight performances'', by which we mean performances without errors, as one of the simplest choices.
Since after notes are defined to be almost definitely played ahead of the succeeding chordal notes or short appoggiaturas, we can divide as $\tilde{\bm\alpha}_I$ and $\tilde{\bm\beta}_I\tilde{\bm y}_I$ if both sub-factors are not empty (otherwise the empty sub-factor is not used for state construction).
If the short appoggiaturas and after notes in the two sub-factors involve only one voice part, and if they do not represent mordents or turns, then they are further divided into factors of intentionally simultaneous notes.
If they involve more than one voice part, there is ambiguity in note ordering across voice parts in general as we explained in Sec.~\ref{sec:OrnamentsAndIndeterminacies}, and they are represented by one bottom-level state.
Note that the possible addition of initial notes and alternations in mordents and turns is incorporated in the state representation.

\begin{figure}[t]
\begin{center}
\includegraphics[clip,width=0.75\columnwidth]{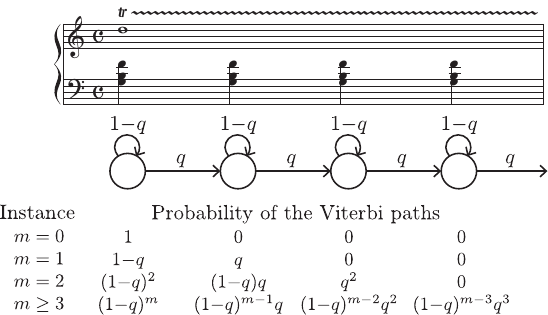}
\end{center}
\caption{Example of a sustained trill with repeated chords. The flow of probability for the Viterbi paths for each state representing a chord and the trill is also shown with the assumption of left-to-right state transitions.}
\label{fig:TrillAndChord}
\end{figure}
However we must make an exception to the above rule since it causes a serious problem for trills and tremolos when they are played in parallel with repeated chords in another voice part.
Example is given in Fig.~\ref{fig:TrillAndChord}.
If we represent each chord with the trill as a state, then these states should have same output probabilities, and particularly, pitch information has no importance in estimating position among these states.
Suppose a straight performance and assume that transition probabilities except for the self transition and transition to the next chord are zero.
The probability of transition to the next state $q$ is nearly equal to the inverse of the number of notes emitted from one state, which indicates $q<1/2$.
Starting with the initial probability of unity at the first state, then, we see that the flow of probability in the Viterbi update does not yield appropriate transition to the second or later state since $1-q>q$ \footnote{The problem of probability flow can be reduced to some extent by using the forward algorithm, but the problem of unreliable estimation still remains.}.
(Note that the IOI information cannot help so much in the presence of a trill or tremolo.)

The problem is significantly reduced if we represent each chord with the trill as two states, one for the attacks of the chordal notes and the other for the subsequent trill notes.
As the simplest possibility, therefore, we represent each $\tilde{\bm\alpha}_I$ or $\tilde{\bm\beta}_I\tilde{\bm y}_I$ by one bottom-level state if the factor contains no trills or tremolos, and otherwise by two bottom-level states.

\begin{figure}[tp]
\begin{center}
\subfigure[The original score, homophonized score, and the corresponding HMM states. The HMM states are illustrated with their state type and main output pitches. The large (resp.~small) smoothed squares indicate top-level (resp.~bottom-level) states.]
{\includegraphics[clip,width=0.95\columnwidth]{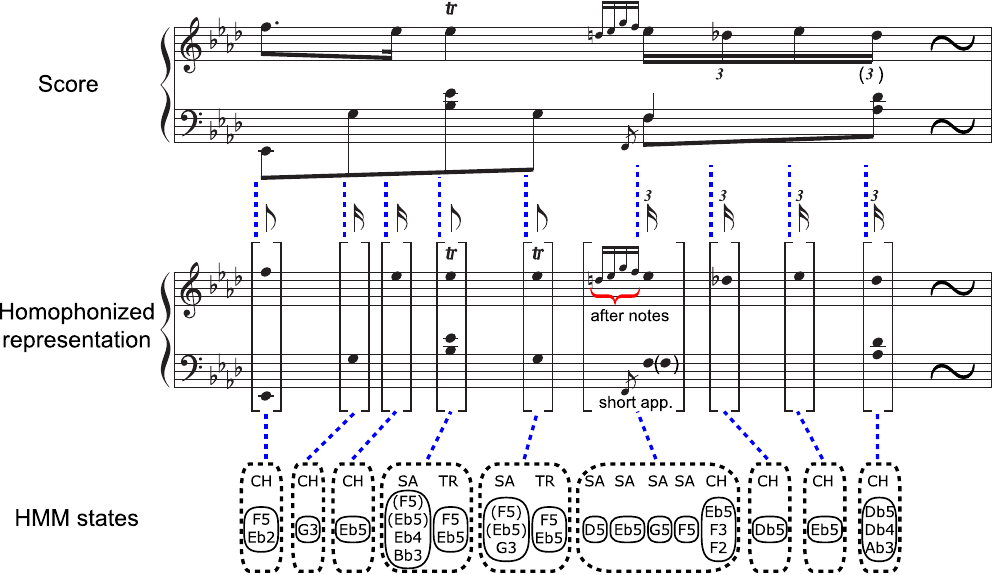}\label{fig:ExHomophonization}}
\subfigure[Representation of the score and its homophonization in terms of Eqs.~(\ref{eq:VoicePart}), (\ref{eq:PolyphonicPassage}), (\ref{eq:HomophonizedSeq}), and (\ref{eq:ComponentsOfHomophonization}). The numerical values indicate score times in units of a quater note, and the symbol $\phi$ denotes ``empty''.]
{\includegraphics[clip,width=0.96\columnwidth]{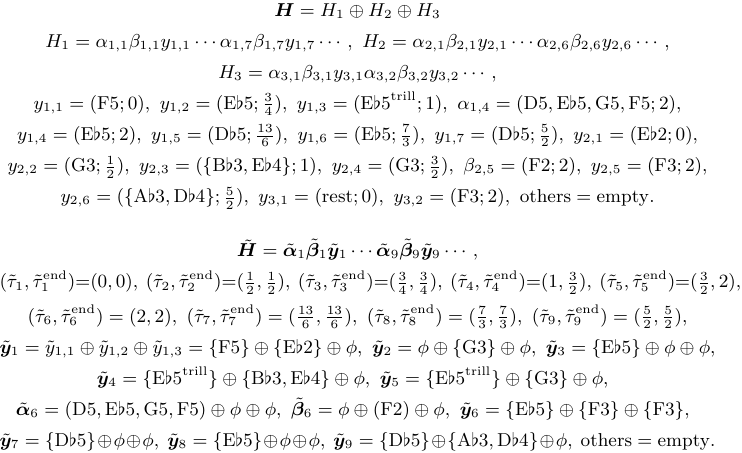}\label{fig:ExScoreRep}}
\end{center}
\caption{Example of homophonization and HMM state construction, together with the score representations.}
\label{fig:ScoreToHMM}
\end{figure}

Three state types are introduced for the bottom HMM.
These are illustrated in Fig.~\ref{fig:ScoreToHMM}, which shows an example of homophonization and HMM state construction for a passage in the solo piano part of Chopin's second piano concerto (second movement).
Type 1 (CH) is used for notes in $\tilde{\bm\beta}_I\tilde{\bm y}_I$ when the factor contains no trills, tremolos, or short appoggiaturas involving multiple voice parts (used in top-level states 1, 2, 3, 6, 7, 8, and 9 in Fig.~\ref{fig:ExHomophonization}).
Type 2 (SA) is used for short appoggiaturas when they involve only one voice part, the chordal notes of $\tilde{\bm\beta}_I\tilde{\bm y}_I$ when it contains trills/tremolos, and after notes (top-level states 4, 5, and 6 in Fig.~\ref{fig:ExHomophonization}).
Finally, Type 3 (TR) is used for the trill/tremolo notes of $\tilde{\bm\beta}_I\tilde{\bm y}_I$ (top-level states 4 and 5 in Fig.~\ref{fig:ExHomophonization}).
Type 1 state is a generalization of a chord state, and it is characterized by an associated metrical note value indicating its duration.
As well as ordinary chordal notes, the state type describes short appoggiaturas in $\tilde{\bm\beta}_I$ and arpeggiated notes in general.
Type 2 state is similar to type 1 state, except that the state is succeeded immediately by another state in a similar sense that a short appoggiatura is succeeded by another note.
Type 3 state describes trills and tremolos in general and is characterized by the continuing emission of notes.

In the following, we describe details of the bottom transition and output probabilities.
Here we suppose that the tempo $v=\Delta t/\Delta \tau$, defined as the ratio of differences of time and score time\footnote{
Tempo defined here is inversely proportional to the conventional one, i.e., beat per minute. It is often used in computational models.}, is given, and its generative model and estimation will be discussed in Sec.~\ref{sec:TempoModel}.
First we explain the transition and output probability for self transition for each state type.

\paragraph{Type 1 or CH}
The self-transition probability $\rho_{\rm CH,CH}$ is determined by matching the expected number of played notes $\sum_{r=1}^\infty r\rho_{\rm CH,CH}^{r-1}(1-\rho_{\rm CH,CH})=1/(1-\rho_{\rm CH,CH})$ to a realistic value $n_e$.
To include the effect of insertions and deletions of notes, $n_e$ is taken as the sum of the number of component notes and a small constant $\epsilon_e$ which represents note insertions and deletions.
We tentatively set $\epsilon_e=0.1$.
The output probability of pitch can be fixed by the distribution of pitches contained in $\tilde{\bm\beta}_I\tilde{\bm y}_I$ for performances without pitch errors, and we can also describe pitch errors by deviations of the distribution \cite{Nakamura2014}.

The output probability of IOI for self transition in the bottom HMM $b_{\rm self}^{\rm IOI}(\delta t)$ can be taken as a mixture of factors as
\begin{equation}\label{eq:SelfIOIProb}
b_{\rm self}^{\rm IOI}(\delta t)=\sum_z\lambda_z b^{\rm IOI}_z(\delta t),
\end{equation}
where $z$ runs through labels ``chord'', ``short app(oggiatura)'', and ``arpeggio'', which corresponds to the distribution of IOI between chordal notes, adjacent short appoggiaturas, and adjacent notes in an arpeggiated chord, respectively.
The $\lambda_z$'s are relative weights that are summed up to one.
The weights are determined by the components in $\tilde{\bm\beta}_I\tilde{\bm y}_I$.
The concrete form of each component distribution of IOI will be explained in Sec.~\ref{sec:Analysis}.

\paragraph{Type 2 or SA}
The self-transition probability $\rho_{\rm SA,SA}$ and the IOI output probability for self transition can be determined in the same way as in the type 1 state.
The pitch distribution is similar as that of type 1 state, but pitches in the trill/tremolo should also be included if the state is succeeded by a type 3 state since they can be performed in between or in precedence to other chordal notes.

\paragraph{Type 3 or TR}
The type 3 state describes a trill or tremolo in general, which is defined by a rapid repetition of multiple chords (typically two) with a total duration indicated with a certain note value.
The bottom-level self-transition probability should thus depend on the expected duration, which is the product of the note value and the tempo, and the expected number of notes per unit time.
Let $\nu_{\rm TR}$ denote the note value of the trill/tremolo, $\bar{n}_{\rm TR}$ the number of notes performed per one repetition, and $\bar{t}_{\rm TR}$ the mean period of the trill/tremolo, and the expected number of emitted notes $n_e$ is given as $n_e=\bar{n}_{\rm TR}v\nu_{\rm TR}/\bar{t}_{\rm TR}$.
Then the self-transition probability $\rho_{\rm TR,TR}$ is given as $n_e+\epsilon_e=1/(1-\rho_{\rm TR,TR})$, as we explained above.
The concrete value of $\bar{t}_{\rm TR}$ is obtained by the performance analysis in Sec.~\ref{sec:Analysis}.

Pitches in the trill/tremolo are used for the pitch distribution, and if addition of after notes is possible, they are also included with small probabilities which can be determined similarly as above.
The IOI distribution for self transition in the bottom HMM is given as a mixture of factors similarly as in Eq.~(\ref{eq:SelfIOIProb}), but now $z$ runs through labels ``chord'' and ``trill''.
The distribution $b_{\rm trill}^{\rm IOI}(\delta t)$ is obtained by analyzing IOIs of trills (see Sec.~\ref{sec:Analysis}).
The relative ratio of $\lambda_{\rm chord}$ and $\lambda_{\rm trill}$ is determined by the constituents of the trill/tremolo.
For example, $\lambda_{\rm chord}=0$ for a one-note trill, $\lambda_{\rm chord}/\lambda_{\rm trill}=1$ for a double trill or a tremolo involving two chords each with two notes, and $\lambda_{\rm chord}/\lambda_{\rm trill}=2$ for a tremolo with two tri-chords.

\paragraph{}
The other transition probabilities in the bottom HMM are determined as follows.
In straight performances, the transition probability to next state is determined by the self-transition probability, and the other probability values are all zero.
Deviations from these values describe performance errors and can be determined by analyzing performance data in principle.
For the lack of sufficient amount of data, however, we set tentative values for these parameters.
For the entering probability, we set $\rho^{(I)}_{{\rm in},k=1}=0.9$ and uniform values for the others $\rho^{(I)}_{{\rm in},k>1}$.
The inter-state probability and exiting probability are set as $\rho^{(I)}_{k,\ell}=0$ if $\ell<k$, and $\rho^{(I)}_{k,{\rm out}}=1-\rho^{(I)}_{k,k}$ if $k$ is the last lower-level state or otherwise $\rho^{(I)}_{k,k+1}=0.9(1-\rho^{(I)}_{k,k})$ and
$\rho^{(I)}_{k,k+2}=\cdots=\rho^{(I)}_{k,{\rm out}}$.

The structure of output probabilities for IOI is a little complicated since it is a Mealy-type output and we are dealing with a hierarchical model.
The output probability for the expanded HMM can be written as $b_{ij}^{\rm IOI}(\delta t)=b_{(I,k)(J,\ell)}^{\rm IOI}(\delta t)$, similarly as the transition probability $a_{ij}$ in Eq.~(\ref{eq:HierarchicalTrProb}).
When $I=J$, we have two transition paths, one for the transition in the bottom HMM and the other for the self transition in the top HMM, corresponding to each term in the right-hand side in Eq.~(\ref{eq:HierarchicalTrProb}), and each path can be associated with an independent IOI distribution.
For the transitions other than self transitions in the bottom HMM, which are immediate transitions, the IOI distribution is modeled by $b^{\rm IOI}_{\rm short~app}(\delta t)$.
The IOI distribution for the path involving the bottom-level transition represents IOIs involving an insertion of events and is written as $b^{\rm IOI}_{II}(\delta t)$, which will be specified in Sec.~\ref{sec:Analysis}.
When $|I-J|$ is large or $|I-J|$ is small and $I>J$, the transition from state $I$ to $J$ describes repeats and skips, and the corresponding IOI distributions are universally represented as a distribution $b^{\rm IOI}_{\rm skip}(\delta t)$.

Finally when $|I-J|$ is small and $I<J$, the transition is a straight transition to the next event or erroneous transitions skipping a few events.
The corresponding IOI can be predicted using the tempo and it is given as 
\begin{equation}\label{eq:OnsetTimePrediction}
\delta t=v(\tilde{\tau}_J-\tilde{\tau}^{\rm end}_{I,k})+({\rm deviation})+({\rm noise}),
\end{equation}
where $\tilde{\tau}_J$ is the score time of the factor $\tilde{\bm\alpha}_J\tilde{\bm\beta}_J\tilde{\bm y}_J$, and $\tilde{\tau}^{\rm end}_{I,k}$ is the score time when the continuation of the corresponding event ends.
The $\tilde{\tau}^{\rm end}_{I,k}$ is same as $\tilde{\tau}_I$ except for the type 3 state, in which case it is $\tilde{\tau}^{\rm end}_I$.
In the above equation, the deviation term adjusts possible deviation, or the ``stolen time'', due to short appoggiaturas and arpeggiated chords, and the noise term describes fluctuations due to motor noise, timing errors, prediction errors, and sudden pauses.
When the factor $(\tilde{\tau}_J-\tilde{\tau}^{\rm end}_{I,k})$ is zero, the transition is immediate and the IOI distribution is modeled by $b^{\rm IOI}_{\rm short~app}(\delta t)$.
Explicit forms and values are described in Sec.~\ref{sec:Analysis}.

A fermata can be represented by a certain enlargement factor of duration and its variance in Eq.~(\ref{eq:OnsetTimePrediction}), etc.
A notated cadenza introduces local deformation of metrical time, and it may be treated as an insertion of the corresponding score time interval.
Although a fermata and a long sequence of grace notes, often written with several note values, are usually indications of a notated cadenza, the distinction with short appoggiaturas requires further information in general.
See discussion in Secs.~\ref{sec:MatchingAlgorithm} and \ref{sec:Evaluation}.

\subsection{Tempo model}\label{sec:TempoModel}

So far we have assumed that the tempo is given in advance.
Since the tempo varies from performance to performance, and it also locally fluctuates during a performance, it is necessary to estimate it continuously for individual performances. For this purpose, we need a tempo model.
Several tempo models and tempo estimation methods have been proposed in Refs.~\cite{Bloch1985,Large1999,Raphael2001,Cemgil2000B,Cemgil2003,Cont2010}.
In the following, we propose a tempo model which describes variation of tempo during performances with erroneous timing as well as expressive timing.
The model is based on that proposed in Refs.~\cite{Raphael2001,Cemgil2000B} with slight modifications.

Variation of tempo is here described as a variation of the local tempo $v_n$, defined as the ratio of IOIs to corresponding note values, i.e.~$v_n=\delta t_n/\nu_n$, where $\delta t_n$ and $\nu_n$ denote duration and note value of the $n$-th note.
(We use $n$, not $m$, to imply that the sequence of local tempos modeled here is not identical to the sequence of all performed notes.)
Since local tempos can only be observed through IOIs, which are subject to noise in human motor controls, a model of their variations should be supplied with such an observational part.

We use a linear dynamical system to model variation of local tempos and their observation through IOIs, following Refs.~\cite{Raphael2001,Cemgil2000B}.
The variation of local tempos are described with a Markov process as
\begin{equation}
v_{n}=v_{n-1}+\frac{\nu_{n-1}v_0}{\nu_{\rm QN}}\epsilon_v,
\end{equation}
where $\nu_{\rm QN}$ is the note value of a quarter note in tick, and $\epsilon_v$ is a stochastic variable with Gaussian distribution with zero mean, which is supposed to be universal for every music piece.
By assuming the tempo variation is globally smooth and scales proportionally with a referential tempo $v_0$, which is taken as the initial tempo, the variation term is proportional to $\nu_{n-1}$ and $v_0$.
Since a universal parameter should be dimensionless, the term is divided by $\nu_{\rm QN}$.
Thus the model is formulated as independent of arbitrary scaling of time and score time in contrast to Refs.~\cite{Raphael2001,Cemgil2000B}.
The standard deviation of $\epsilon_v$ is denoted by $\sigma_v=\sqrt{\langle\epsilon_v^2\rangle}$.

The observation of IOI is modeled as
\begin{equation}\label{eq:TempoModelIOIOutput}
\delta t_n=\nu_nv_{n}+e_t.
\end{equation}
Here $e_t$ represents a noise term resulting from fluctuating onset times.
In musical performances including those during practice, onset time is subject to erroneous timing, which results from errors in rhythm and added pauses, in addition to noise from motor controls.
We can represent these two different causes in the observed IOI as a mixture of two noise sources as
\begin{equation}
e_t=\xi_1 \epsilon_t^{(1)}+\xi_2 \epsilon_t^{(2)},
\end{equation}
where $\epsilon_t^{(1)}$ and $\epsilon_t^{(2)}$ represent noise sources due to motor controls and erroneous timing, and $\xi_1$ and $\xi_2$ represent relative weights, satisfying $\xi_1+\xi_2=1$.
Phenomenologically, the distribution of erroneous timing includes large values that are more properly approximated by a widespread distribution such as the Cauchy distribution than the Gaussian.
For efficient inference, however, Gaussian approximation is more convenient and we can indeed use the switching Kalman filter \cite{Kim1994}.
Thus we will assume that $\epsilon_t^{(1)}$ and $\epsilon_t^{(2)}$ are Gaussians and their standard deviations, $\sigma_t^{(1)}$ and $\sigma_t^{(2)}$, and the weight are determined in Sec.~\ref{sec:Analysis}.

%%%%%%%%%%%%%%%%%%%%%%%%%%%%%%%%%%%%%%%%%%%%%%
\section{Analysis and model parameters}\label{sec:Analysis}
%%%%%%%%%%%%%%%%%%%%%%%%%%%%%%%%%%%%%%%%%%%%%%

\subsection{Performance preparation}\label{sec:PerformanceData}

For the purpose of analyzing performances to fix details of the model, and of evaluating the score-performance matching algorithms described in later sections, we prepared piano performance data of several musical pieces by several performers.
Scores are prepared in the MusicXML format and notes in the performance data, which are recorded in MIDI files, are matched to notes in the score by hand.
When matched notes in the score could not be found or there were ambiguities, they are labeled as ``unmatched notes'' with possible candidate matched score notes.

We recorded performances of three pianists, two conservatory students in piano and one amateur player, for musical pieces in which ornaments are extensively used.
The performances were recorded during practices and they contain relatively many performance errors, repeats, and skips.
The pieces were chosen to efficiently cover a wide range of ornamental figures in the common practice period.
They are the first harpsichord part of Couperin's Allemande \`a deux clavecins (the first piece of the ninth ordre in second book of pi\`eces de clavecin), the solo piano part in the second movement of Beethoven's first piano concerto, the third movement of Beethoven's second piano concerto, and the second movement of Chopin's second piano concerto.
The Couperin's piece contains many mordents and turns in a manner typical of the Baroque period.
The second movement of Beethoven's first concerto contains long sustained trills with bass passages in other voice parts as well as other short ornaments.
The third movement of his second concerto contains many short appoggiaturas.
The movement of Chopin's concerto contains many arpeggios, trills, after notes, and short appoggiaturas intertwined in polyphony, together with many polyrhythmic passages and his habitual coloratura-like passages.
The slow movements were also intentionally chosen to analyze and test temporally complex passages.

\subsection{IOI distributions}\label{sec:IOIDistribution}

\begin{figure}[tbp]
\begin{center}
\subfigure[Chord]
{\includegraphics[clip,width=0.475\columnwidth]{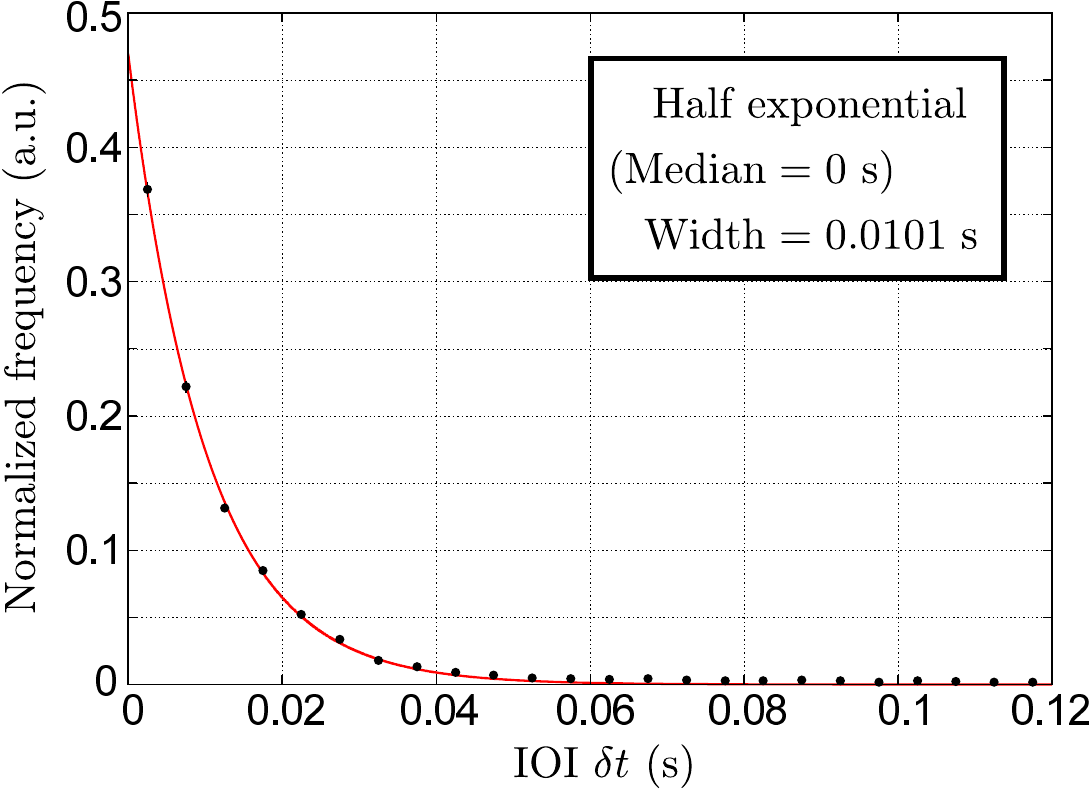}}
\subfigure[Trill]
{\includegraphics[clip,width=0.475\columnwidth]{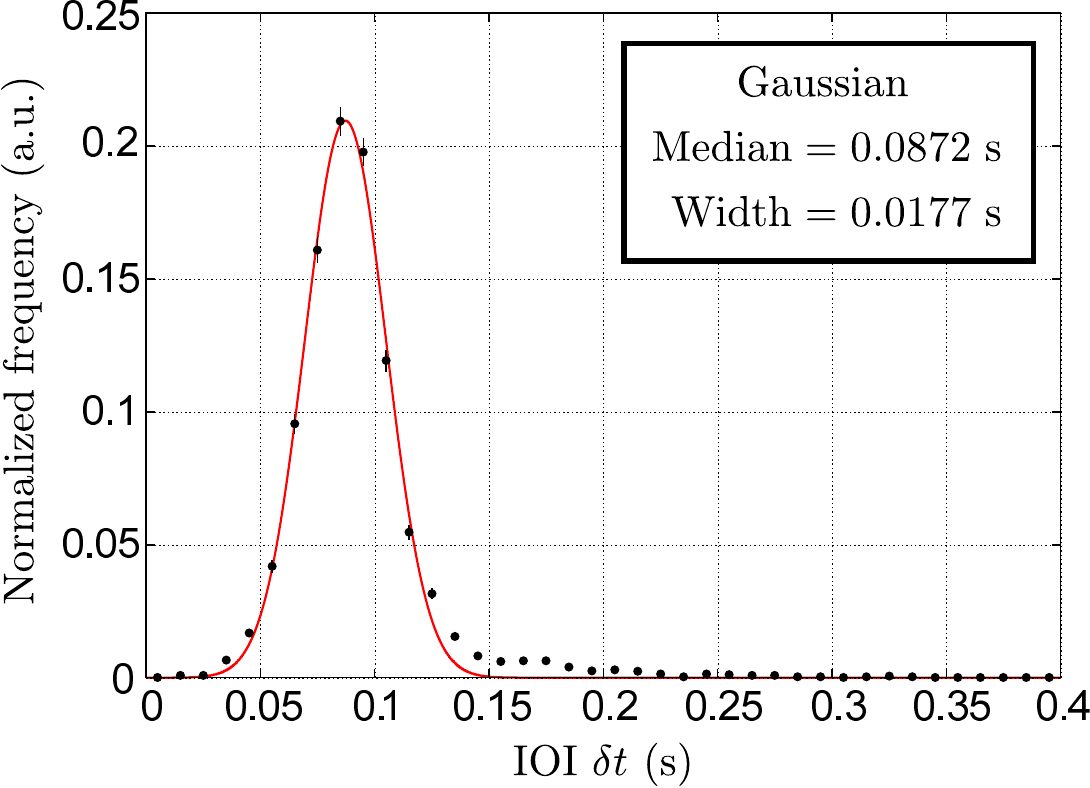}}\\
\subfigure[Short appoggiatura]
{\includegraphics[clip,width=0.475\columnwidth]{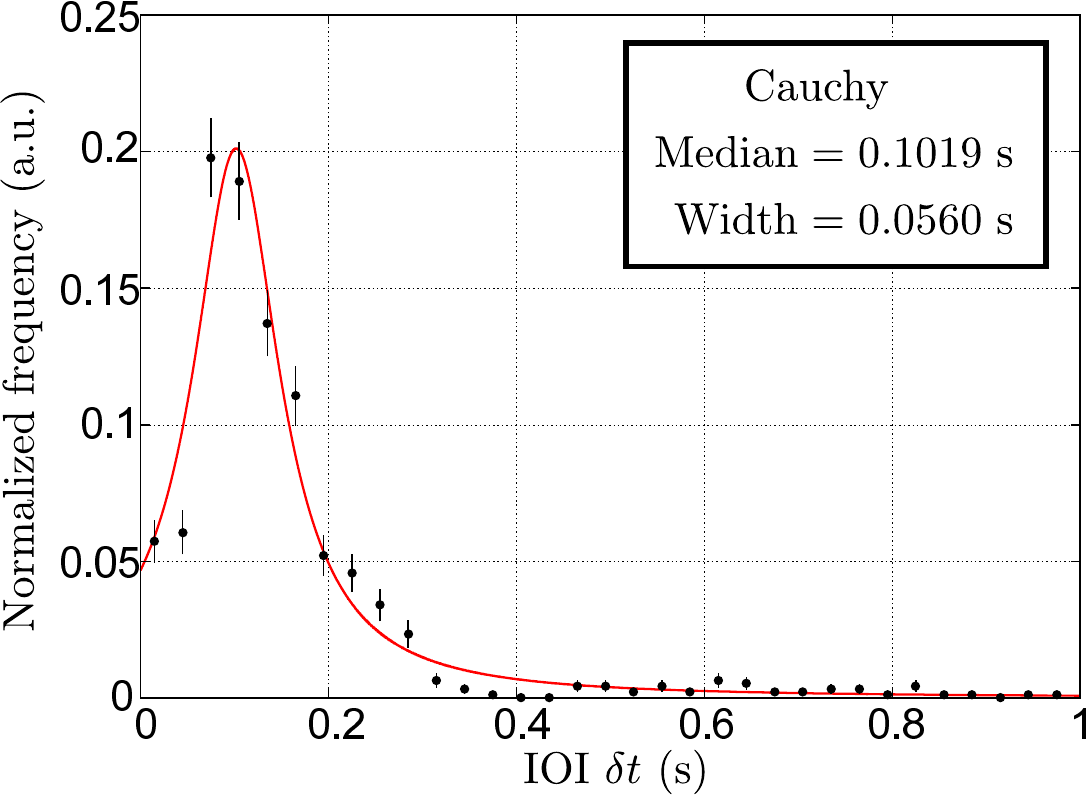}}
\subfigure[Arpeggio]
{\includegraphics[clip,width=0.475\columnwidth]{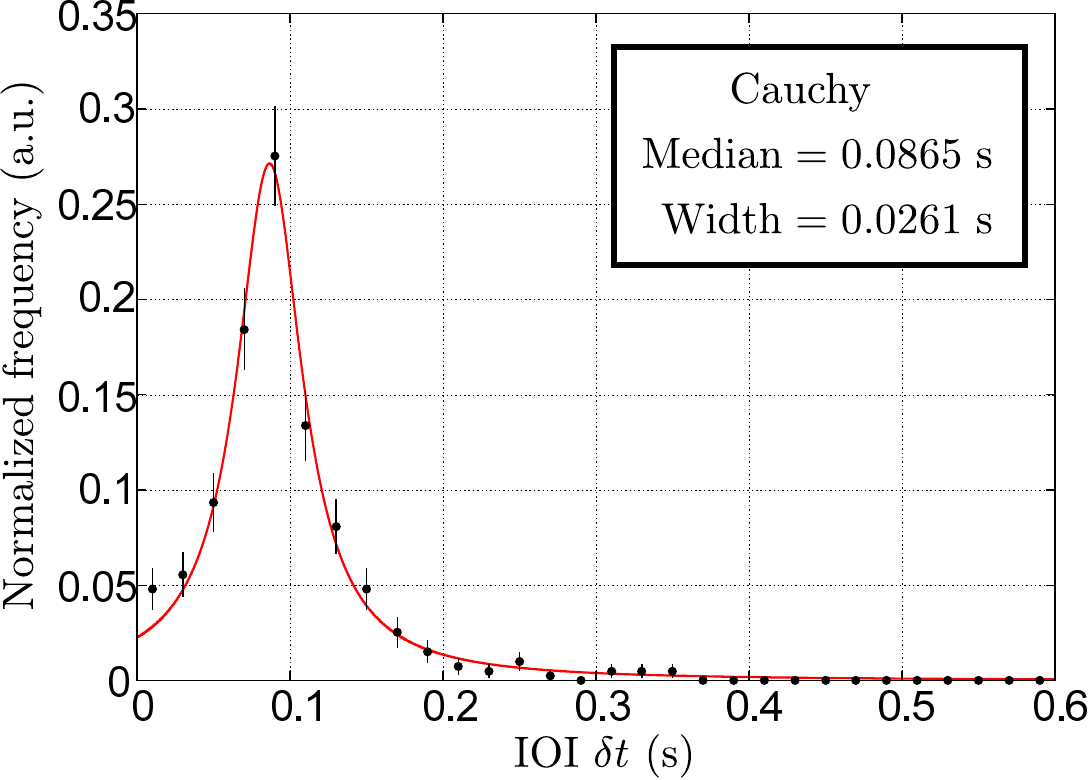}}\\
\subfigure[Repeat/skip/insertion]
{\includegraphics[clip,width=0.475\columnwidth]{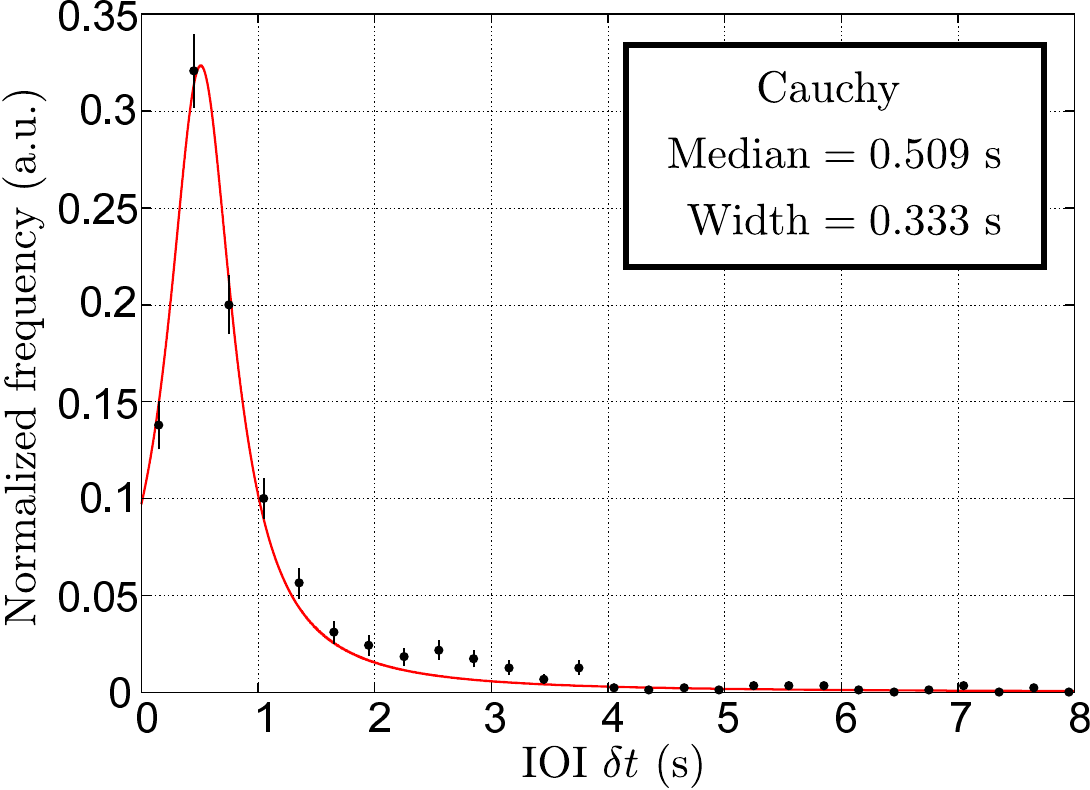}\label{fig:IOIRepeatSkip}}
\subfigure[All self-transition IOI distributions]
{\includegraphics[clip,width=0.475\columnwidth]{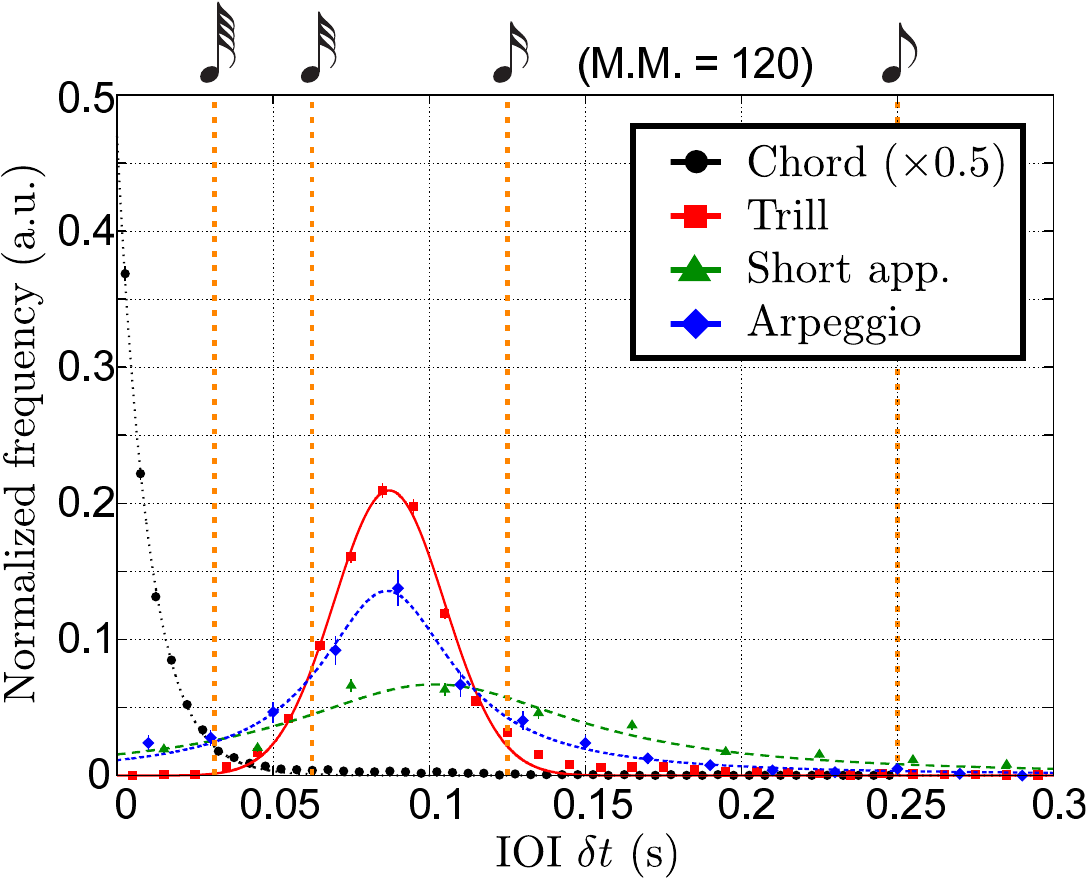}\label{fig:IOIAll}}
\end{center}
\caption{Distributions of IOIs. The width here means the standard deviation (resp.~rate parameter, half width at half maximum) for the Gaussian (resp.~half exponential, Cauchy) distribution.}
\label{fig:IOIDistribution}
\end{figure}
Distributions of IOIs of notes in chords, trills, short appoggiaturas (including after notes), and arpeggios are shown in Fig.~\ref{fig:IOIDistribution}, together with fitted distribution functions.
The distribution of IOI involving repeats, skips, and insertions of chords, taken from the performance data in Ref.~\cite{Nakamura2014} is also shown.
Because it was hard to determine the functional form a priori for most of the distributions, we tested the Gaussian, exponential, and Cauchy distribution for each and selected the best fitted one in terms of $R^2$.
The fitted distributions and values of the parameters are also shown in the figure.

The clean exponential distribution of chord IOI indicates the onsets of chordal notes obey a Poisson process approximately.
For chords and trills, the IOI distributions have tails in larger values that cannot be well described by the Gaussian or exponential function.
They mostly result from erroneous actions that cannot be described by one simple distribution.
For example, a small peak around $\delta t=0.17$ can be explained by deletions of a trill note, which result in IOIs about twice as large as normal IOIs with a central value of $\delta t\simeq0.87$.
Since these contributions are not dominating in frequency, they are tentatively represented as a mixed component of a Cauchy distribution, which is taken as the distribution depicted in Fig.~\ref{fig:IOIRepeatSkip}.
The other distributions have more complicated forms, presumably from complex origins, but we do not pursue the precise identification of them.

All of the self-transition IOI distributions are normalized and shown in Fig.~\ref{fig:IOIAll} for a small IOI range, together with referential durations of several note values in tempo ${\rm M.M.}=120$.
One can check that the distributions for ornaments overlap significantly with the range of inter-chord IOIs and IOIs of chordal notes, which indicates that simple threshold methods do not work very well for clustering performed notes into musical events including ornaments, and thus it is important to use explicit temporal information.
We see that the distributions for short appoggiatura and arpeggio have non-vanishing values near zero IOI, and they are more widespread than the distribution for trill.
Some of the features seem to be generic empirically, but a fully extensive analysis is needed to draw conclusions on generality, piece dependencies, and performer dependencies.
Further studies may have importance in understanding expressive music performance.

The above results yield a concrete form for $b^{\rm IOI}_{\rm chord}$, $b^{\rm IOI}_{\rm short~app}$, $b^{\rm IOI}_{\rm arpeggio}$, $b^{\rm IOI}_{\rm trill}$, and $b^{\rm IOI}_{\rm skip}$.
We also use the same form of $b^{\rm IOI}_{\rm skip}$ for the distribution for insertion of events $b^{\rm IOI}_{II}$.

\subsection{Tempo model and onset time prediction}

The tempo estimation and the onset time prediction, which is encoded in Eq.~(\ref{eq:OnsetTimePrediction}), are related.
The independent parameters of the tempo model are $\sigma_v$, $\sigma_t^{(1)}$, $\sigma_t^{(2)}$, and $\xi_1$ (Sec.~\ref{sec:TempoModel}).
The estimated tempo by the switching Kalman filter is independent of a scale factor of the variances, and we fix the absolute values of the variances by the value of $\sigma_t^{(1)}$, which is set as 0.014 s by matching the variance to that of the IOI distribution of chordal notes with the assumption that these have the common source of noise.

The other parameters can be determined by minimizing the prediction error of onset time.
To do this efficiently, we first determine $\sigma_v$ in the simple model with $\xi_1=1$ with the data of expressive piano performances \cite{Hashida2008}, in which rhythmic errors and unexpected pauses are rare.
Next we optimize $\sigma_t^{(2)}$ and $\xi_1$ with the performance data described in Sec.~\ref{sec:PerformanceData}.
The obtained values are $\sigma_t^{(2)}=0.16$ s, $\sigma_v=0.03$, and $\xi_1=0.95$.
Note that the optimal values can depend on music piece and performance's tendencies (e.g.,~whether it is concert-ready or during practice), and therefore, there is room for further adaptation.

\begin{figure}[tbp]
\begin{center}
{\includegraphics[clip,width=0.55\columnwidth]{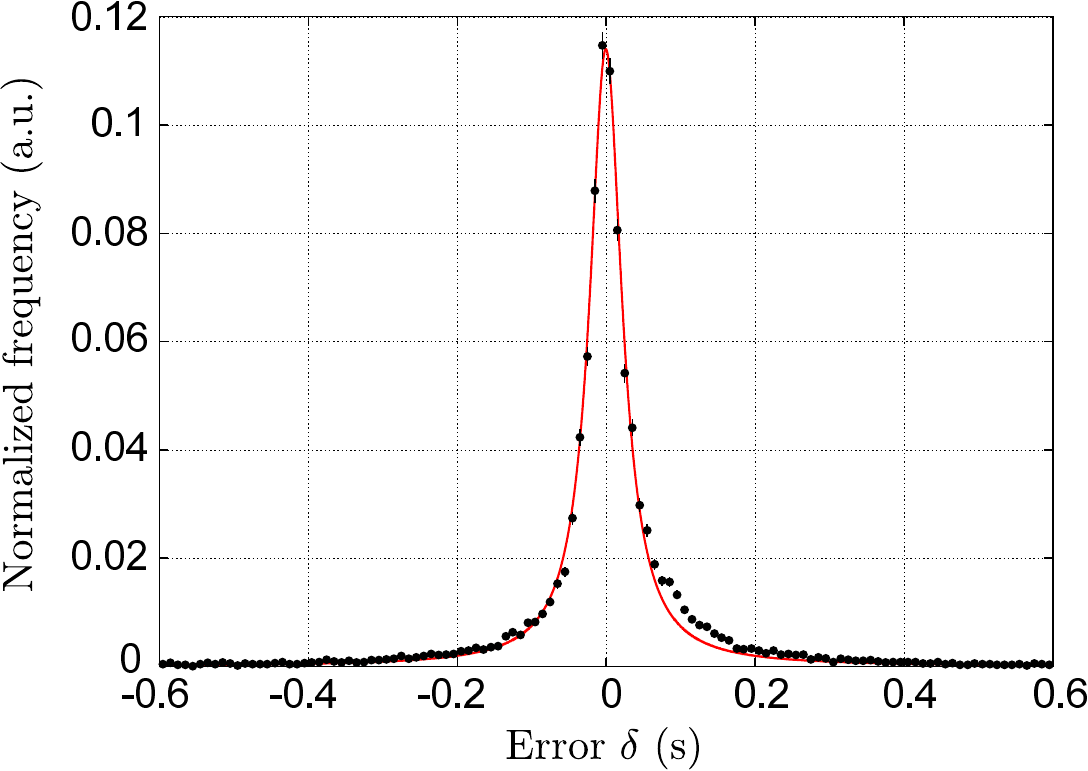}}
\end{center}
\caption{Distribution of prediction error of next onset time.}
\label{fig:OnsetPredictionError}
\end{figure}
Using the result in Sec.~\ref{sec:IOIDistribution}, expected timing of short appoggiaturas and arpeggios can be used for the deviation term in Eq.~(\ref{eq:OnsetTimePrediction}).
The form of the noise term in the equation can be obtained from the distribution of predicted error of next onset notes (Fig.~\ref{fig:OnsetPredictionError}).
The Cauchy distribution was chosen as a well-fitted one (${\rm Width}=0.0264$ s, ${\rm Median}<0.001$ s).
Assuming a mixed source of error as in Eq.~(\ref{eq:TempoModelIOIOutput}), the Cauchy-like distribution may arise from averaging over all note values and tempos, even if each term is a Gaussian noise.
We see some excess on the positive side ($70$ ms $\lesssim\delta\lesssim 200$ ms), indicating an unidentified source of delayed timing.
Possible causes are agogic and caesura, and short delays due to difficulty of preparing performance actions.

%%%%%%%%%%%%%%%%%%%%%%%%%%%%%%%%%%%%%%%%%%%%%%
\section{Application to score-performance matching}\label{sec:Matching}
%%%%%%%%%%%%%%%%%%%%%%%%%%%%%%%%%%%%%%%%%%%%%%

\subsection{Description of algorithms}\label{sec:MatchingAlgorithm}

Given the stochastic generative model of performance, the score-performance matching problem can be restated as an inference problem of hidden state sequence $(i_m)_{m}$ given observations of performed notes $(p_m,t_m)_{m}$, which can be efficiently solved by the Viterbi algorithm.
As stated in Ref.~\cite{Nakamura2014}, both online and offline matching algorithms can be constructed in a similar manner, by using a partial sequence $(p_{m'},t_{m'})_{m'=1}^{m}$ of past observations and a full sequence $(p_{m'},t_{m'})_{m'=1}^M$ including future observations, respectively, for calculating the most likely sequence of states.

When dealing with temporal information, tempo must be estimated simultaneously with score positions.
One can consider a two-dimensional state space of score position and tempo and perform a joint estimation \cite{Otsuka2011,Montecchio2011,Duan2011}.
Generally, inferences using Monte Carlo methods or discretization methods are then necessary since the search space is large.
However, such inferences are inefficient for the current model because it allows arbitrary repeats and skips and the search space is much larger.

We instead consider coupling the score-position model and the tempo model by updating the probabilities of the two models alternately, similarly as in Ref.~\cite{Cont2010}.
Given an estimated value of tempo, we can update the $m$-th factor in Eq.~(\ref{eq:PerformanceProbLatter}), which is given as a product of Eq.~(\ref{eq:HierarchicalTrProb}) and output probabilities of pitch and IOI described in Sec.~\ref{sec:EventModel}.
Using the result of estimated score positions, estimation of the current tempo can be done with the switching Kalman filter as mentioned in Sec.~\ref{sec:TempoModel}, whose result is used for the next update of the probability of score positions.
For algorithmic details of the switching Kalman filter, see Ref.~\cite{Kim1994}.

The estimation of tempo has to be done carefully for performances with ornaments since it is not meaningful to update tempo for every note in ornaments, which does not have a definitive metrical note value.
To solve the problem, the tempo is updated only when the estimated state in the top HMM is updated.
To avoid estimation errors due to stolen time in arpeggios, short appoggiaturas, etc., the time interval between the detected times of states, instead of that of between note onsets, is used as the observable of the tempo model.

Because we allow arbitrary repeats and skips, the computational complexity for score-position estimation is large in the conventional Viterbi algorithm.
To reduce the computational complexity, we use the uniform skip model in Ref.~\cite{Nakamura2014}, assuming uniform probability for large repeats and skips.
The processing time for one Viterbi update increases proportionally to the number of HMM states.
It was less than 2 ms for the solo piano part of the second movement of Chopin's second concerto for which the number of states was 1354, with 2 GHz Intel Core i7 CPU and 8 GB RAM.

We have explained the performance model to construct the matching algorithms in the previous sections.
To improve the matching accuracy, however, we use a slightly modified model and parameter values.
First parameter to change is the probability of large repeats and skips $\bar{\gamma}$.
A certain period of time is necessary to identify the correct score position after repeats and skips.
We can use a probability value smaller than the realistic value to prevent estimated score positions from jumping to an erroneous score position.
The estimation results then tend to stay within the neighborhood of the current score position, which has the effect of stabilizing the estimation robust against local performance errors.
We can also increase the width $\Delta$ of the predicted IOI in Eq.~(\ref{eq:OnsetTimePrediction}) from the realistic value in Fig.~\ref{fig:OnsetPredictionError} to allow noise and deviations in time as long as it will not destroy the whole estimation result.
We optimized these parameters by data and the obtained values were $(\bar{\gamma},\Delta)=(e^{-20},0.4~{\rm s})$ for the online algorithm and $(e^{-40},0.3~{\rm s})$ for the offline algorithm.
Since the fitted Cauchy distribution in Fig.~\ref{fig:IOIRepeatSkip} cannot well describe the steep decrease in the small IOI range, which is relatively important because of overlap with other IOI distributions in Fig.~\ref{fig:IOIAll}, we set almost zero probability below a threshold of $0.3$ s.

Temporal fluctuation is expected to be large at fermatas and notated cadenzas, and one can introduce further enlargement of the width of the predicted IOI at these events.
However, we empirically found, with a small number of examples, that the matching algorithms work well at these events without such labor.
Thus they are treated in the same way as ordinary events.

\subsection{Comparison with other algorithms}\label{sec:Comparison}

Here we compare our algorithms with other algorithms, and discuss their features, advantages, and disadvantages.
The most unique feature of our algorithms is that they can handle arbitrary repeats and skips, as it is a direct extension of the algorithms in Ref.~\cite{Nakamura2014}.
Particularly there have been no other offline matching algorithms that can cope with large repeats and skips, and there have been no online matching algorithms that handle ornaments in addition.
Generally our algorithms based on a stochastic performance model are best characterized by their robustness against many performance errors and ornamentations, since their treatment requires a complex set of rules which are encoded in the performance model but otherwise relatively hard to keep consistency and efficiency.
As we encode the indeterminacies of ornaments listed in Table \ref{tab:ListIndeterminacies} in the model, the derived algorithms in effect hold rules to treat them as much as the model can tell.
We next discuss detailed comparison separately for online and offline algorithms.

For online matching, the algorithm for MIDI performance with explicit treatment of ornaments in Ref.~\cite{Dannenberg1988} is the only one in the literature.
As explained in Sec.~\ref{sec:SignificanceForMatching}, the preprocessing method in Ref.~\cite{Dannenberg1988} often cause troubles under frequent performance errors, repeats and skips, which is avoided by our stochastic method.
For example, a proper activation of the preprocessor can fail with an erroneously deleted note just before a trill or with a direct skip to a trill event, which can induce further troubles in matching succeeding events.
Our algorithm handles these possibilities by associated probability values in the model and the algorithm can recover quickly even when some estimation error happens.
Since the algorithm in Ref.~\cite{Dannenberg1988} treats temporal information in a manner similar to a threshold method, the preprocessing and the treatment of short appoggiaturas or arpeggios can fail as we discussed in Sec.~\ref{sec:IOIDistribution}.
In our algorithm, tendencies in the temporal structure are more precisely described by IOI distributions in the temporal HMM, and therefore, a better estimation can be made.
Finally, to extend the preprocessing method to the case of polyphonic passages, for example, when an arpeggio is played with a trill, seems to be not easy because of the above arguments.
Our algorithm is applicable to any scores in the form of Eq.~(\ref{eq:PolyphonicPassage}).

Treatment of ornaments in offline matching is discussed in Ref.~\cite{Gingras2011}, where both pitch and IOI informations are used to detect ornamental notes.
In their technique, ornaments are treated in a general way and indeed they handle both notated and free ornaments in a unified way.
In our algorithm, the pitch and IOI informations are used through the performance model and the way they are used is optimized in the sense of maximal likelihood.
Free ornaments can be treated as note insertions in principle, and this may work well for simple ornaments such as mordents and arpeggios.
A proper treatment of heavy free ornamentation may require additional refinement because the algorithm uses only local timing by the Markovian assumption and this can be distorted by heavy ornamentation.
The voice information used in Refs.~\cite{Heijink2000,Gingras2011}, which is important when voice asynchrony influences the ordering of notes across voices, is not incorporated in the present algorithm.
An advantage of our algorithm is computational efficiency.
Compared to the computation time reported in Ref.~\cite{Gingras2011}, our algorithm is likely faster, but a rigorous comparison has not been conducted.

It is a big challenge to derive or identify explicit rules that work for particular situations, which is generally hard and a potential drawback of the stochastic method.
A thorough quantitative analysis of the algorithms is also difficult because it is hard at the moment to perform a deeper analysis of stochastic models like HMMs due to their complex sequential nature.
We perform quantitative evaluations of the algorithms in the next section, and we encourage further studies on these problems and leave them for the future.

%%%%%%%%%%%%%%%%%%%%%%%%%%%%%%%%%%%%%%%%%%%%%%
\section{Evaluation of the score-performance matching algorithms}\label{sec:Evaluation}
%%%%%%%%%%%%%%%%%%%%%%%%%%%%%%%%%%%%%%%%%%%%%%

\subsection{Online matching (score following)}\label{sec:onlineMatchingEvaluation}

We evaluated effectiveness of the algorithms with the error rate of score-performance matching compared to the hand-matched results.
The same performance data explained in Sec.~\ref{sec:PerformanceData} was used.
The combination of upper mordent and turn in the Couperin's piece was replaced by a direct turn in the prepared score.
For comparison, we implemented two other matching algorithms in addition to the one in Sec.~\ref{sec:MatchingAlgorithm}.
One is based on the preprocessing method proposed in Ref.~\cite{Dannenberg1988}.
To handle repeats and skips, we combine the preprocessor to the basic HMM-based algorithm without modeling ornaments.
In case of successive trills, which is not described in Ref.~\cite{Dannenberg1988}, the preprocessor is kept in the trill state while the score position moves to the next when the condition of exiting the state is satisfied.
The other one is also based on an HMM without modeling ornaments, but with referring to an explicit realization of scores.
The information of ornaments was given through the corresponding labels of the realized notes.
For example, a mordent, a turn, or a trill is expressed as a set of explicitly realized notes with identical labels.
Except for preprocessing in the former algorithm and threshold for clustering chords, which was taken as 35 ms for both of the algorithms as in Ref.~\cite{Nakamura2014}, the temporal information is not used in these algorithms.

\begin{table}[tb]
\caption{Error rates (\%) of the online matching algorithms. Error rate of the note-level (resp.~score-time-level) matching is indicated outside (resp.~inside) parentheses. Pieces indicate those described in Sec.~\ref{sec:PerformanceData}.
}
\label{tab:EROnline}
\begin{center}
\begin{tabular}{ccccc}\hline\hline
\shortstack{Piece\\~}&\shortstack{\# of \\onsets}&\shortstack{HMM w/\\ornaments}&\shortstack{HMM w/o\\ornaments}&\shortstack{Preprocessing\\~}\\
\hline
Couperin&1763&4.71 (3.74)&16.5 (13.5)&24.2 (23.5)\\
Beethoven No.1&17587&3.28 (3.22)&7.83 (7.74)&28.2 (28.1)\\
Beethoven No.2&5861&2.73 (2.70)&5.49 (4.62)&8.36 (8.33)\\
Chopin&16241&10.4 (9.61)&16.2 (14.6)&28.2 (25.7)\\
\hline\hline
\end{tabular}
\end{center}
\end{table}
The results of the online matching are shown in Table \ref{tab:EROnline}.
There, we indicated the error rates of precise matching of performed notes to score notes, referred to as note-level matching, and those of less precise matching where performed notes are matched only to score time without considering the matching to individual score notes, referred to as score-time-level matching.
We see that the algorithm based on the temporal HMM with ornaments yielded the lowest error rate, the one based on the HMM without modeling ornaments yielded the second lowest, and the one with preprocessing had the worst error rate in every case.
It is confirmed that the explicit modeling of ornaments is indeed effective.

The relatively high error rates of the algorithm with preprocessing for Beethoven's first concerto and Chopin's concerto are mainly due to accidental troubles in the preprocessing of trills caused by performance errors and failure of score-position estimation.
They are severe especially for passages with several succeeding trills and those with a sustaining trill with other voice parts as expected.
The latter case is often problematic for the algorithm without modeling ornaments, demonstrating the fact that it is hard to correctly match all notes by treating the indeterminacies of trill notes simply as deletions and insertions without particular tuning of model parameters for trills.
For both algorithms without modeling ornaments, threshold clustering of notes often fail in passages with short appoggiaturas and arpeggios, as seen in the relatively high error rates for Couperin's and Chopin's pieces.

A contribution of error rate from estimation errors in a certain period of time after repeats and skips before identifying the correct resumption score position is unavoidable \cite{Nakamura2014}, which becomes manifest by comparing the error rates with the results for offline alignment in the next subsection.
It is a large portion of error rates for the algorithm with the HMM with ornaments.
Another source of typical estimation errors of the algorithm is after notes following trills which are misidentified as the trill notes.
The ambiguity in distinguishing the after notes from trill notes with pitch errors can be reduced if probability of erroneous pitches is low, or if the durational information of the trill is used, for example with the variable duration model.
Occasionally a problem occurs in detecting a note following a trill if the pitch of the note is same as the trill notes.
Precise modeling of duration of the trill can help, but the problem in principle can only be solved by relaxing the strict real-time online condition and use some kind of delayed decision \cite{Dannenberg1988}.

Another major situation of estimation errors is polyrhythmic passages, coloratura-like passages, and fast passages involving both hands, as manifested mostly in the results of Chopin's piece.
In such passages, effect of voice asynchrony influences ordering of notes across voices \cite{Heijink2000}.
Similar case is for passages with a slow turn or a long chain of short appoggiaturas/after notes which is superposed with other voice parts, for which the degree of overlaps of the ornamental notes with the other voice part is variable and uncertain.
Although such reordering of notes can be treated as performance errors in principle, appropriate treatment such as tuning model parameters is difficult without voice information.

Some fermatas and notated cadenzas appeared in Beethoven's first concerto and Chopin's concerto.
We confirmed that the algorithm with the temporal HMM works well at these events even without special treatment, as mentioned in Sec.~\ref{sec:MatchingAlgorithm}.
This is because of relatively large width of the noise term in Eq.~(\ref{eq:OnsetTimePrediction}), indicating that the algorithm is also robust against tempo rubato.
In general we can apply online adaptation of the parameter for further improvement.

\subsection{Offline matching (alignment)}

\begin{table}[t]
\caption{Comparison of errors for the algorithm with HMM with ornaments and the matchers in Ref.~\cite{Gingras2011} (3-step matcher) and Ref.~\cite{Heijink2000} (structure matcher). The integers outside (resp.~inside) the parentheses indicate the number of discrepancies between the hand-matched results given by Gingras and McAdams \cite{Gingras2011} (resp.~by the present authors Nakamura et al.) and the estimated results. ``Etude'' refers to the Etude in C minor, Op. 10, No. 12; ``Fantaisie'' refers to the Fantaisie Impromptu, Op. 66 (both by Fryderyk Chopin). The performances were distributed on floppy discs from Yamaha Music Corp. The percentages show the averaged error rates for each piece.}
\label{tab:ERAlignComparison}
\begin{center}
\begin{tabular}{llcccc}\hline\hline
\shortstack{Piece\\~}&\shortstack{Performance\\(disc no./track)}&\shortstack{\# of \\onsets}&\shortstack{HMM w/\\ornaments}&\shortstack{3-step\\matcher \cite{Gingras2011}}&\shortstack{Structure\\matcher \cite{Heijink2000}}\\
\hline
Etude&YMM 900148/12&465&1 (1)&0 (0)&0 (0)\\
Etude&YMM 900202/2&466&1 (0)&0 (0)&0 (0)\\
Etude&YPA 1069E/1&466&1 (1)&0 (0)&1 (1)\\
Etude&YPA 1070E/27&462&1 (2) &1 (2) &2 (1)\\
Etude&YPA 1100E/8&465&0 (0)&0 (0)&0 (0)\\
%Etude&{\bf Total}&2324&4 (4) [0.17 (0.17\%)]&1 (2) [0.04 (0.08\%)]&3 (2) [0.13 (0.08\%)]\\\hline
Etude&{\bf Total}&2324&4 (4)&1 (2)&3 (2)\\
&&&[0.17 (0.17)\%]&[0.04 (0.08)\%]&[0.13 (0.08)\%]\\\hline
Fantaisie&YPA 1077E/3&244&19 (19)&0 (0)&0 (0)\\
Fantaisie&YPA 1100E/5&247&19 (17)&0 (6)&0 (6)\\
%Fantaisie&{\bf Total}&491&38 (36) [7.74 (7.33)\%] &0 (6) [0 (1.22)\%] &0 (6) [0 (1.22)\%]\\
Fantaisie&{\bf Total}&491&38 (36) &0 (6) &0 (6)\\
&&&[7.74 (7.33)\%]&[0 (1.22)\%]&[0 (1.22)\%]\\
\hline\hline
\end{tabular}
\end{center}
\end{table}
Error rate of offline score-performance matching is also evaluated.
First we compared our algorithm by the HMM with ornaments to the state-of-the-art algorithms in Refs.~\cite{Heijink2000,Gingras2011}, for which the pieces (excerpts from Chopin's Etude Op.~10-12 and Fantaisie Impromptu) and performance data in Ref.~\cite{Gingras2011} is used.
The performances include no large repeats or skips and errors were rare (under about 1\% of the total performed notes), except that there were reordered notes across voice parts due to voice asynchrony especially in performances of Fantaisie Impromptu.
Since the preprocessing method cannot directly be applied for offline matching and the pieces did not contain any ornaments, the algorithm with preprocessing and the one with the HMM without modeling ornaments were not used for the comparison.
The results are listed in Table \ref{tab:ERAlignComparison}, where the number of discrepancies between the hand-matched results, both by the authors in Ref.~\cite{Gingras2011} and by the current authors, is shown.

We see that there were slightly more (although a few) errors with our algorithm compared to the other algorithms for the performances of the Etude.
We confirmed that the main cause of additional errors with our algorithm was reordered notes across voice parts.
There were more errors for the performances of Fantasie Impromptu with our algorithm, whereas other algorithms remain very accurate.
We also confirmed that most of the errors with our algorithm were around reordered notes across voice parts and thus induced by voice asynchrony, indicating the fact that the use of voice information is important in the situation.
It is worth mentioning that the hand-matched results by different annotators differed slightly mostly because of the ambiguity in interpreting performance errors (it is also mentioned in Ref.~\cite{Gingras2011}).
For the second performance of Fantasie Impromptu (Disc: YPA 1100E, Track: 5), for example, there were six discrepancies between the hand matched results by Gingras and McAdams \cite{Gingras2011} and by the present authors.
Four of them involved pitch error (in our interpretation) and two involved notes reordered across hands.

\begin{table}[t]
\caption{Error rates (\%) of the offline matching algorithms (note-level matching). See the caption of Table \ref{tab:EROnline}.
}
\label{tab:EROffline}
\begin{center}
\begin{tabular}{cccc}\hline\hline
\shortstack{Piece\\~}&\shortstack{\# of \\onsets}&\shortstack{HMM w/\\ornaments}&\shortstack{HMM w/o\\ornaments}\\
\hline
Couperin&1763&2.67&12.1\\
Beethoven No.1&17587&1.41&5.86\\
Beethoven No.2&5861&0.87&3.16\\
Chopin&16241&6.96&11.2\\
\hline\hline
\end{tabular}
\end{center}
\end{table}
Next, we evaluate the algorithms with the same pieces used in Sec.~\ref{sec:onlineMatchingEvaluation}.
Since the performance data include large repeats and skips, we used the algorithms by the temporal HMM with and without modeling ornaments.
The results show again that the explicit modeling of ornaments is effective (Table \ref{tab:EROffline}).
Errors due to repeats and skips and after notes following trills are much reduced in the offline matching because it is usually necessary to observe several performed notes to correctly detect these events and inference from the future is effective, which is incorporated in the backtracking step in the Viterbi algorithm.
A major source of estimation error is now voice asynchrony and heavy local distortion of performance by frequent errors.
The former elucidates again the importance of voice information for further improvement.
There were also mismatches of global score positions in pieces with repeated sections of same or similar passages. Such mismatches would remain in principle in the presence of large repeats and skips, but they may be reduced by using prior knowledge on the tendencies of repeats and skips \cite{Nakamura2014}.

%%%%%%%%%%%%%%%%%%%%%%%%%%%%%%%%%%%%%%%%%%%%%%
\section{Summary and discussion}\label{sec:Conclusion}
%%%%%%%%%%%%%%%%%%%%%%%%%%%%%%%%%%%%%%%%%%%%%%

In this paper, we proposed an HMM-based performance model incorporating indeterminacies in realization of ornaments.
The indeterminacies of ornaments are represented in terms of pitch and temporal informations with several state types, which we formalized for quite general polyphonic passages.
Our model describes the temporal information with one dimension in the state space, and we showed that it is equivalent to an HMM with IOI output.
The model accommodates tempo variation, performance errors, arbitrary repeats, and skips while keeping computational efficiency for inference, and has advantages in music information processing.

We carried out an analysis on piano performances, obtained phenomenological fitting functions for IOI distributions involving ornaments, and determined their parameters.
From the performance analysis, we quantitatively found that the IOI distributions for trills, short appoggiaturas, and arpeggios have significant overlaps with that of chordal notes and that for inter-chord events.
It is also found that the IOI distributions for short appoggiatura and arpeggio have non-vanishing values near zero IOI, and they are more widespread than that for trill.
The results motivate further extensive analyses.

We applied the model to score following and offline score-performance matching, and obtained computationally efficient and highly accurate algorithms that can handle performance errors, ornaments, and arbitrary repeats and skips.
We confirmed that the explicit modeling of ornaments indeed works effectively and the online algorithm is more robust against errors and unexpected repeats and skips than the preprocessing method.
A major cause of estimation errors is the reordering of notes across voices due to voice asynchrony and widely stretched ornaments, and the result suggests that refinements such as incorporating voice information are necessary for an essential solution.
It would be necessary to relax the assumed strict temporal structure in homophonization and model loosely coupled voice parts.
Since the refined model would probably require more computational cost for inference, we expect that the current model will remain profitable as a computationally efficient model that yields accurate score-performance matching for most scores and performances.
How to incorporate the voice information \cite{Heijink2000,Gingras2011} while keeping computational efficiency and capability of handling repeats and skips is an open problem.

Another possible application of the model is music transcription and rhythm quantization.
Specifically such algorithms can be constructed by equipping the model with a score/language model \cite{Otsuki2002}, and intelligent transcription algorithms that can handle ornaments may be obtained.
Similarly, automatic extraction of ornaments in performances of unknown score may be possible to some extent.
In general, recognition of ornaments from performance has ambiguity in principle since several score representations are possible as any other recognition problems, and the model can provide a measure of naturalness in terms of probability.
The idea of using several different performances for transcribing a score may also have importance.
Our matching algorithm can also be applied to prepare large-scale corpus of performances for performance analysis and automatic rendering of performance.

As we have stressed several times, to incorporate the voice structure into the stochastic performance model is one of the most important steps to go forward.
Since polyphony is a general and prominent character of music, studying asynchrony and inter-dependency between voices has importance in music information and interests in music research in general.
A thorough analysis of the model is also important to understand its validity and limitation in music information processing quantitatively.

%%%%%%%%%%%%%%%%%%%%%%%%%%%%%%%%%%%%%%%%%%%%%%
\section*{Acknowledgements}
%%%%%%%%%%%%%%%%%%%%%%%%%%%%%%%%%%%%%%%%%%%%%%

The authors are grateful to Ayumu Yamanaka and Tadayuki Hayasaka for useful discussions and their cooperation in piano performance, and Bruno Gingras for providing the evaluation data and giving many helpful suggestions for revisions in the text.
The author E.N. wishes to thank Yasuyuki Saito and Tomohiko Nakamura for fruitful discussions.
This work is supported in part by Grant-in-Aid for Scientific Research from Japan Society for the Promotion of Science, Nos.~23240021 (S.S. and N.O.), 25880029 and 15K16054 (E.N.).

\appendix

%%%%%%%%%%%%%%%%%%%%%%%%%%%%%%%%%%%%%%%%%%%%%%
\section{Details of homophonization}\label{sec:Homophonization}
%%%%%%%%%%%%%%%%%%%%%%%%%%%%%%%%%%%%%%%%%%%%%%

We first convert upper and lower mordents and turns with short appoggiaturas and after notes, according to the tentative representations in Fig.~\ref{fig:TentativeRepresentations}, and glissandos with their explicit realizations.
Notated cadenzas are also converted to sequences of chords and inserted in Eq.~(\ref{eq:PolyphonicPassage}), with enlargement of rests or chords in other voice parts at a proper score position usually indicated with a fermata.
The score representation after these manipulations can be written in the same as in Eq.~(\ref{eq:PolyphonicPassage}), and we reuse the symbol $\bm H$ for it.
Let $\alpha_{v,i}$, $\beta_{v,i}$, and $y_{v,i}$ denote the $i$-th factor in the $v$-th voice part of a score $\bm H$ in Eq.~(\ref{eq:PolyphonicPassage}).
We define the trill part of a factor $y_{v,i}$ as
\begin{equation}
{\rm TR}(y_{v,i})=
\begin{cases}
y_{v,i},&\text{if $y_{v,i}$ is a tremolo};\\
\{\text{trills in $y_{v,i}$}\}, &\text{if $y_{v,i}$ contains a trill};\\
\emptyset, &{\rm otherwise}.
\end{cases}
\end{equation}
A factor $y_{v,i}$ is said purely trill-like if $y_{v,i}={\rm TR}(y_{v,i})$.
We assume that there is no redundancies in every $H_v$, that is, no two succeeding factors $\alpha_{v,i}\beta_{v,i}y_{v,i}$ are both empty, nor both $\alpha_{v,i}\beta_{v,i}$ are empty and both $y_{v,i}$ are purely trill-like and identical.
There is no loss of generality because if there are such factors we can concatenate them to reduce $H_v$ to an equivalent voice part, and any $H_v$ can be reduced to a voice part without redundancies after a finite number of such reductions.

Now let $\tau_{v,i}$ denote the score time of $\alpha_{v,i}\beta_{v,i}y_{v,i}$, and we have $\tau_{v,i}<\tau_{v,i+1}$ for all $i$.
We can construct a sequence $(\tilde{\tau}_I)_I$ of score times such that $\tilde{\tau}_I<\tilde{\tau}_{I+1}$ for all $I$, and for all $v$ and $i$ we can find an $I$ s.t.~$\tilde{\tau}_I=\tau_{v,i}$ and for all $I$ we can find some $v$ and $i$ s.t.~$\tilde{\tau}_I=\tau_{v,i}$.
Then a tentative homophonization $\tilde{\bm H}'$ of $\bm H$ is constructed as
\begin{align}
\tilde{\bm H}'&=\prod_I \tilde{\bm\alpha}'_I\tilde{\bm\beta}'_I\tilde{\bm y}'_I,\\
\tilde{\bm\alpha}'_I&=\bigoplus_v\tilde{\alpha}'_{I,v},\quad
\tilde{\bm\beta}'_I=\bigoplus_v\tilde{\beta}'_{I,v},\quad
\tilde{\bm y}'_I=\bigoplus_v\tilde{y}'_{I,v},
\end{align}
where
\begin{align}
\tilde{\alpha}'_{I,v}&=\begin{cases}
\alpha_{v,i},&{\rm if}~\tilde{\tau}_I=\tau_{v,i};\\
\emptyset,&{\rm otherwise},
\end{cases}\\
\tilde{\beta}'_{I,v}&=\begin{cases}
\beta_{v,i},&{\rm if}~\tilde{\tau}_I=\tau_{v,i};\\
\emptyset,&{\rm otherwise},
\end{cases}\\
\tilde{y}'_{I,v}&=\begin{cases}
y_{v,i},&{\rm if}~\tilde{\tau}_I=\tau_{v,i};\\
{\rm TR}(y_{v,i}),&{\rm if}~\tau_{v,i}<\tilde{\tau}_I<\tau_{v,i+1},
\end{cases}
\end{align}
We define $\tilde{\tau}^{\rm end}_I=\tilde{\tau}_I$ if the factor $\tilde{\bm\alpha}'_I\tilde{\bm\beta}'_I\tilde{\bm y}'_I$ contains no trills or tremolos, and $\tilde{\tau}^{\rm end}_I=\tilde{\tau}_{I+1}$ otherwise.

The final stage to construct the homophonization $\tilde{\bm H}$ is to remove redundancies in $\tilde{\bm H}'$ since we are interested only in note onsets.
For this, we concatenate an empty factor $\tilde{\bm\alpha}'_I\tilde{\bm\beta}'_I\tilde{\bm y}'_I$ to the previous $(I-1)$-th factor and keep $\tilde{\tau}^{\rm end}_{I-1}$ is unchanged.
When $\tilde{\bm\alpha}'_I\tilde{\bm\beta}'_I$ is empty and $\tilde{\bm y}'_I$ is purely trill-like and is included in the $(I-1)$-th factor ($\tilde{\bm y}'_I\subset\tilde{\bm y}'_{I-1}$), then $I$-th factor is deleted and we put $\tilde{\tau}^{\rm end}_{I-1}=\tilde{\tau}^{\rm end}_{I}$.
This process will end in finite steps and then we obtain $\tilde{\bm H}$.

\end{document}